%% file: paper.tex
\documentclass[twoside,leqno,twocolumn]{article}  
\usepackage{ltexpprt}
\usepackage{wrapfig}
\usepackage{amsmath,amssymb, graphicx}
\usepackage[ruled,linesnumbered]{algorithm2e}
\usepackage{algorithmic}
\usepackage{mdframed}
\usepackage{paralist} % also compact lists
\usepackage{color}
\usepackage{multirow}
\usepackage{rotating}
\usepackage{amsmath}
\usepackage{hyperref}
\usepackage[mathscr]{eucal} %% For \mathscr
\usepackage{amsbsy} %% For \boldsymbol
\usepackage{subfigure}
\usepackage{booktabs}
\usepackage{xcolor}
\usepackage{tikz}
\usetikzlibrary{snakes}
\usepackage{multirow}
\usepackage{epstopdf}
\usepackage{balance}
\usepackage{tablefootnote}
\usepackage{empheq} % added by christos - makes nice box for equations
\usepackage{moresize}
\usepackage{array}
\usepackage{enumitem}
\usepackage{sidecap}
\setlist{nolistsep}

\newcommand{\tensor}[1]{\underline{ \mathbf{#1} }}

\usepackage[font=footnotesize,labelfont=bf,skip=0pt]{caption}

\DeclareCaptionType{copyrightbox}
\usepackage{url}

\newcounter{ALC@tempcntr}% Temporary counter for storage
\input{dfn}

\def\x{{\mathbf x}}

\newcommand{\method}{\textsc{OCTen}\xspace}

\newcommand{\codeurl}{\url{ http://www.cs.ucr.edu/~egujr001/ucr/madlab/src/OCTen.zip}}

\begin{document}

\author{
Ekta Gujral\\
       UC Riverside\\
       {egujr001@ucr.edu}
\and
Ravdeep Pasricha \\
 UC Riverside\\
 {rpasr001@ucr.edu}
 \and
 Tianxiong Yang\\
  UC Riverside\\
  {tyang022@ucr.edu}
  \and
Evangelos  E. Papalexakis\\
       UC Riverside\\
       {epapalex@cs.ucr.edu}
}

\date{}
\title{\method: Online Compression-based Tensor Decomposition}
%\title{Triple-Sparse, Parallel, and Incremental Tensor Decomposition}
\maketitle
\input{000abstract}
\input{010introduction}
\input{022preliminaries}
\input{020problem}

\input{030method}
\input{040experiments}
\input{050related}
\input{060conclusions}

\hide{
\section{Acknowledgments}
{\scriptsize
Research was supported by the Department of the Navy, Naval Engineering Education Consortium under Award no. N00174-17-1-0005, by the National Science Foundation Grant no. EAGER 1746031, and an Adobe Data Science Research Faculty Award. Any opinions, findings, and conclusions or recommendations expressed in this material are those of the author(s) and do not necessarily reflect the views of the funding parties. We would also like to thank Xia Ben Hu for fruitful discussions on the problem.

}
}

% References should be produced using the bibtex program from suitable
% BiBTeX files (here: strings, refs, manuals). The IEEEbib.bst bibliography
% style file from IEEE produces unsorted bibliography list.
% -------------------------------------------------------------------------
\balance
\bibliographystyle{abbrv}
\bibliography{BIB/refs}
\newpage
\balance
\input{070Supplement}
\end{document}

%% file: dfn.tex
\newcommand{\hide}[1]{}
\newcommand{\ben}{\begin{enumerate*}}
\newcommand{\een}{\end{enumerate*}}
\newcommand{\bit}{\begin{itemize*}}
\newcommand{\eit}{\end{itemize*}}

%% file: 000abstract.tex
\begin{abstract}
Tensor decompositions are powerful tools for large data analytics as they jointly model multiple aspects of data into one framework and enable the discovery of the latent structures and higher-order correlations within the data. One of the most widely studied and used decompositions, especially in data mining and machine learning, is the Canonical Polyadic or CP decomposition. However, today's datasets are not static and these datasets often dynamically growing and changing with time. To operate on such large data, we present \method the first ever compression-based online parallel implementation for the CP decomposition.We conduct an extensive empirical analysis of the algorithms in terms of fitness, memory used and CPU time, and in order to demonstrate the compression and scalability of the method, we apply \method to big tensor data. Indicatively, \method performs on-par or better than state-of-the-art online and offline methods in terms of decomposition accuracy and efficiency, while saving up to \text{\em {40-200 \%}} memory space.		
\end{abstract}

%% file: 010introduction.tex
\section{Introduction}
\label{sec:intro}
A Tensor is a multi-way array of elements that represents higher-order or multi-aspect data. In recent years, tensor decompositions have gained increasing popularity in big data analytics \cite{papalexakis2016tensors}. In higher-order structure, tensor decomposition are capable of finding complex patterns and  higher-order correlations within the data. \hide{Much like}Corresponding to matrix factorization tools like SVD (Singular Value Decomposition), there exist generalizations for the tensor domain, with the most widely used being CANDECOMP/PARAFAC or CP \cite{PARAFAC} which extracts interpretable factors from tensor data, and Tucker decomposition \cite{tucker3}, which is known for estimating the joint subspaces of tensor data. In this work we focus only on the CP decomposition, which is extremely effective in exploratory knowledge discovery on multi-aspect data.
In the era of information explosion, data is generated or modified in large volume. In such  environments, data may be \hide{increased or decreased in any of its dimensions with high velocity.}added or removed from any of the dimensions with high velocity. When using tensors to represent this dynamically changing data, an instance of the problem is that of a ``streaming'', ``incremental'', or ``online'' tensors\footnote{Notice that the literature (and thereby this paper) uses the above  terms as well as ``dynamic'' interchangeably.}. Considering an example of time evolving social network interactions, where a large number of users interact with each other every second (Facebook users update $\approx 684K$ information and Twitter users send $\approx 100K$ tweets every single minute\footnote{http://mashable.com/2012/06/22/data-created-every-minute/}); every such snapshot of interactions is a new incoming slice(s) to the tensor on its ``time'' mode, which is seen as a streaming update. Additionally, the tensor may be growing in all of its n-modes, especially in complex and evolving environments such as online social networks. In this paper, our goal is, given an already computed CP decomposition, to {\em track} the CP decomposition of an online tensor, as it receives streaming updates, 1) {\em efficiently},  being much faster than re-computing the entire decomposition from scratch after every update, and utilizing small amount of memory, and 2) {\em accurately}, incurring an approximation error that is as close as possible to the decomposition of the full tensor. For exposition purposes, we focus on the  time-evolving/streaming scenario, where a three-mode tensor grows on the third (``time'') mode, however, our work extends to cases where more than one modes is online.

\begin{figure}[!ht]
	\begin{center}
		\includegraphics[clip,trim=0.1cm 3.0cm 0.5cm 3.0cm,width = 0.45\textwidth]{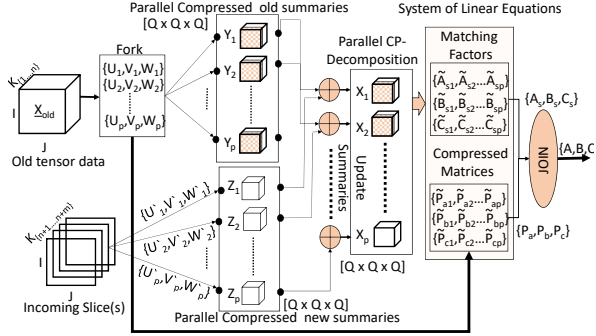}
		\caption{\bf{\method framework. Compressed tensor summaries $\tensor{Y_p}$ and $\tensor{Z_p}$  are obtained by applying randomly generated compression matrices $(\mathbf{U}_p, \mathbf{V}_p,\mathbf{W}_p)$ and $(\mathbf{U}_p^{'}, \mathbf{V}_p^{'},\mathbf{W}_p^{'})$ to $\tensor{X}_{old}$ and $\tensor{X}_{new}$ or incoming slice(s) respectively. The updated summaries are computed by $\tensor{X}_p$ = $\tensor{Y}_p+\tensor{Z}_p$. Each $\tensor{X}_p$ is independently decomposed in parallel. The update step anchors all compression and factor matrices to a single reference i.e. $(\mathbf{P}_a, \mathbf{P}_b,\mathbf{P}_c)$ and $(\mathbf{A}_s,\mathbf{B}_s,\mathbf{C}_s)$, and solves a linear equation for the overall A, B, and C.} }
		\label{fig:crown_jewel}
	\end{center}
	\vspace{-0.4in}
\end{figure}
%Tensor decompositions are inevitable tool for analyzing such multi-way or multi-aspect data. However, traditional tensor decomposition algorithms, like TUCKER and CANDECOMP/PARAFAC(CP), did not perform well on online tensors, due to the need to process whole data and more space and time requirements.  
As the volume and velocity of data grow, the need for time- and space-efficient online tensor decomposition is imperative. There already exists a modest amount of prior work in online tensor decomposition both for Tucker \cite{austin2016parallel,SunITA} and CP \cite{nion2009adaptive,zhou2016accelerating}.However, most of the existing online methods \cite{austin2016parallel,zhou2016accelerating,nion2009adaptive} \hide{We *may* wanna say here ``with the exception of SamBaTen'' but maybe not, because we may distract from the main point.}, model the data in the full space, which can become very memory taxing as the size of the data grows. There exist memory efficient tensor decompositions, indicatively MET for Tucker \cite{kolda2008scalable} and PARACOMP \cite{sidiropoulos2014parallel} for CP, neither of which are able to handle online tensors. In this paper, we fill \hide{exactly}that gap.
 
Online tensor decomposition is a challenging task due to the following reasons. First, maintaining high-accuracy (competitive to decomposing the full tensor) using significantly fewer computations and memory than the full decomposition calls for innovative and, ideally, sub-linear approaches. Second, operating on the full ambient space of \hide{the}data, as the tensor is being updated online, leads to super-linear increase in time and space complexity, rendering such approaches hard to scale, and calling for efficient methods that work on memory spaces \hide{that}which are significantly smaller than the original ambient data dimensions. Third, in many real settings, more than one modes of the tensor may receive streaming updates at different points in time, and devising a flexible algorithm that can handle such updates seamlessly is a challenge.
To handle the above challenges, in this paper, we propose to explore how to decompose online or incremental tensors based on CP decomposition. We specifically study: (1) How to  make parallel update method based on CP decomposition for online tensors? (2) How to identify latent component effectively and accurately after decomposition? Answering the above questions, we propose \method (Online Compression-based Tensor Decomposition) framework. Our contributions are summarized as follows:
\begin{itemize}
	\item {\bf Novel Parallel Algorithm} We introduce \method, a novel compression-based algorithm for online tensor decomposotion that admits an efficient parallel implementation. We do not limit to 3-mode tensors, our algorithm can easily handle higher-order tensor decompositions. 
	\item {\bf Correctness guarantees} By virtue of using random compression, \method can guarantee the {\em identifiability} of the underlying CP decomposition in the presence of streaming updates.
	\item {\bf Extensive Evaluation} Through experimental evaluation on various datasets, we show that \method  provides stable decompositions (with quality on par with state-of-the-art), while offering up to \text{\em {40-250 \%}} memory space savings.
\end{itemize}

{\bf Reproducibility}: We make our Matlab implementation publicly available at link \footnote{\codeurl}. Furthermore, all the small size datasets we use for evaluation are publicly available on same link.
\hide{
The rest of paper is organized as follows. Section \ref{sec:prelim} presents the preliminaries of the tensor and its decompositions. Section \ref{sec:problem} formally describes problem and Section \ref{sec:method} presents our proposed method for the scalable online tensor decompositions. After presenting the experimental results in Section \ref{sec:experiments}, we discuss related works in Section  \ref{sec:related}. Then we conclude in Section \ref{sec:conclusions}.
}

%% file: 022preliminaries.tex
\section{Preliminaries}
\label{sec:prelim}
\textbf{Tensor} : A tensor is a higher order generalization of a matrix. An $N$-mode tensor is essentially indexed by $N$ variables. In particular, regular matrix with two variables i.e. $I$ and $J$ is 2-mode tensor, whereas data cube ( $I$,$J$ and $K$) is viewed as a 3-mode tensor $\tensor{X} \in \mathbb{R}^{I\times J \times K} $.
The number of modes is also called "order". Table \ref{table:t2} contains the symbols used throughout the paper. For the purposes of background explanation, we refer the reader to \cite{papalexakis2016tensors} for the definitions of Khatri-Rao and Kronecker products which are not necessary for understanding the basic derivation of our framework.
\hide{A tensor is a higher order generalization of a matrix. In order to avoid overloading the term ``dimension'', we call an $I\times J \times K$ tensor a three ``mode'' tensor, where ``modes'' are the numbers of indices used to index the tensor. The number of modes is also called ``order''. Table \ref{table:t2} contains the symbols used throughout the paper. In the interest of space, we also refer the reader to \cite{papalexakis2016tensors} for the definitions of Kronecker and Khatri-Rao products which are not necessary for following the basic derivation of our method. }
\hide{
\textbf{Slice} :  A slice is a (N-1)-dimension partition of tensor $\tensor{X}$ where size is varied only in temporal mode and other non-temporal modes remain fixed.\\ \hide{The tensor $\tensor{X}$ can be sliced as \hide{There are three categories of slices : }horizontal ($\tensor{X}$(i,:,:)) , lateral ($\tensor{X}$(:,j,:)), and frontal ($\tensor{X}$(:,:,k)) mode. }

}
% %***************************Table***************************
\hide{\begin{table}[h!]
	\vspace{-0.2in}
\ssmall
\begin{center}
 {\rowcolors{1}{gray!30}{white!100!yellow!10}
\begin{tabular}{  c c c c  }
\hline
\textbf{Notation} &\textbf{Example}&& \textbf{Definition} \\ 
\hline
\hline
Underline boldface capital letters&$\tensor{X}$ && Tensor \\ 
\hline
Boldface capital letters&$ \mathbf{X} $ && Matrix  \\ 
\hline
Boldface lowercase letters&$\mathbf{x}$ && Column vector \\ 
\hline
Lowercase letters&$x$ && Scalar \\ 
\hline
-&$\mathbb{R}$ && Set of Real Numbers  \\ 
\hline
Small circle&$\circ$ && Outer product  \\ 
\hline
-&$\lVert \mathbf{A} \rVert_F, \| \mathbf{a} \|_2$&& Frobenius norm, $\ell_2$ norm \\
\hline 
Index&$\mathbf{x}(i),\mathbf{X}(i,j),\tensor{X}(i,j,k)$ && Element of vector, matrix, or tensor \\ 
\hline
O-sum&$ \oplus  $&&  Summation (\cite{papalexakis2016tensors})\\
\hline
O-times and O-dot&$\otimes, \odot$ && Kronecker and Khatri-Rao product (\cite{papalexakis2016tensors})\\
\end{tabular}}
 \caption{\bf{Table of symbols and their description}}
\label{table:t2}
\end{center}
\vspace{-0.2in}
\end{table}}
%%%%%%%%%%%%%%%%%%%%%%%%%%%%%%%%%%%%%
% %***************************Table***************************
\begin{table}[h!]
	\small
	\begin{center}
				\begin{tabular}{ c|c }
				\hline
				Symbols & Definition \\ 
				\hline
				\hline
				$\tensor{X},\mathbf{X}, \mathbf{x},x$ & Tensor, Matrix, Column vector, Scalar \\ 
				\hline
				$\mathbb{R}$ & Set of Real Numbers  \\ 
				\hline
				$\circ$ & Outer product  \\ 
				\hline
				$\lVert \mathbf{A} \rVert_F, \| \mathbf{a} \|_2$& Frobenius norm, $\ell_2$ norm \\
				\hline 
				$\oplus$&  Summation \\ 
				\hline
				$\otimes $& Kronecker product\cite{papalexakis2016tensors}\\
				\hline
				$\odot$ & Khatri-Rao product\cite{papalexakis2016tensors}\\
				\hline
		\end{tabular}
		\caption{\bf{Table of symbols and their description}}
		\label{table:t2}
	\end{center}
	%\vspace{-0.2in}
\end{table}
%%%%%%%%%%%%%%%%%%%%%%%%%%%%
\vspace{-0.2in}

\textbf{Canonical Polyadic Decomposition}: One of the most popular and extensively used tensor decompositions is the Canonical Polyadic (CP) or CANDECOMP/ PARAFAC decomposition \cite{carroll1970analysis,PARAFAC,bader2015matlab} referred as CP decomposition. Given a N-mode tensor $\tensor{X}$ of dimension $\mathbb{R}^{I_1 \times I_2\times \dots I_N}$, its CP decomposition can be written as $\mathbf{A}^{(n)} \in \mathbb{R}^{I_n \times R}$, where $n=(1,2,\dots N)$ and R represents the number of latent factors or upper bound rank on tensor $\tensor{X}$. Henceforth, the 3-mode tensor of size $\mathbb{R}^{I  \times J\times K}$ can be represented as a sum of rank-one tensors:
$
\tensor{X} \thickapprox \sum_{r=1}^R \mathbf{A}(:,r) \circ \mathbf{B}(:,r)\circ \mathbf{C}(:,r)
$
where $\mathbf{A}\in  \mathbb{R}^{I\times R}, \mathbf{B}\in \mathbb{R}^{J \times R},  \mathbf{C}\in \mathbb{R}^{K\times R}$. The unfold n-mode tensor $\tensor{X}$ can be written as khari-rao product of its modes as 
$
\tensor{X}^{(n)} \thickapprox   \mathbf{A}^{(n)} (\mathbf{A}^{(N)} \dots \odot \mathbf{A}^{(n+1)} \odot \mathbf{A}^{(n-1)} \odot \mathbf{A}^{(1)})^T  
$. 
\hide{To decompose the tensor we minimize the loss function as:
$\mathcal{L}  \approx \min||\tensor{X}^{(n)} -   \mathbf{A}^{(n)} (\mathbf{A}^{(K)} \dots \odot \mathbf{A}^{(k+1)} \odot \mathbf{A}^{(k-1)} \odot \mathbf{A}^{(1)})^T||_F^2
$.} We refer the interested reader to several well-known surveys that provide more details on tensor decompositions and its applications \cite{kolda2009tensor,papalexakis2016tensors}.

%% file: 020problem.tex
\section{Problem Formulation}
\label{sec:problem}
In many real-world applications, data grow dynamically and may do so in many modes. For example, given a dynamic tensor in a movie-based recommendation system, organized as \text{\em{users $\times$ movie $\times$ rating $\times$ hours}}, the number of registered users, movies watched or rated, and hours may all increase over time. Another example is network monitoring sensor data where tons of information like source and target IP address, users, ports etc., is collected every second. This nature of data gives rise to update existing decompositions on the fly or online and we call it incremental decomposition.  In such conditions, the  update needs to process the new data very quickly, which makes non-incremental methods to fall short because they need to recompute the decomposition for the full dataset.
The problem that we solve is the following:

\begin{mdframed}[linecolor=red!60!black,backgroundcolor=gray!20,linewidth=2pt,    topline=false,rightline=false, leftline=false]

{\bf Given} (a) an existing set of {\em summaries} \{$\tensor{Y}_{1},\tensor{Y}_{2} \dots \tensor{Y}_{p}$\}, which approximate tensor $\tensor{X}_{old}$ of size \{ $ I^{(1)} \times I^{(2)} \times \dots I^{(N-1)} \times  t_{old}$\} at time \textit{t} , (b) new incoming batch of slice(s) in form of tensor $\tensor{X}_{new}$ of size \{$ I^{(1)} \times I^{(2)} \times \dots I^{(N-1)} \times  t_{new}$\}, find updates of ($\mathbf{A}^{(1)},\mathbf{A}^{(2)}$ , \dots , $\mathbf{A}^{(N-1)}$, $\mathbf{A}^{(N)}$) {\bf incrementally} to approximate tensor $\tensor{X}$ of dimension \{$ I^{(1)} \times I^{(2)} \times \dots I^{(N-1)} \times I^{(N)}$\} and rank R, where $ I^{(N)} = (t_{old}+t_{new})= I_{1\dots n}^{(N)} + I_{(n+1)\dots m}^{(N)}$ after appending new slice or tensor to $N^{th}$ mode.
 
\end{mdframed}

%% file: 030method.tex
\section{\method Framework}
\label{sec:method}
As we mention in the previous section, to the best of our knowledge, there is no algorithm in the literature that is able to efficiently compress and incrementally update the CP decomposition in the presence of incoming tensor slices. However, there exists a method for static data \cite{sidiropoulos2014parallel}. Since this method considers the tensor in its entirety, it cannot handle streaming data and as the data size grows its efficiency may degrade.\hide{, since it handles the full data in one shot.} In this section, we introduce \method, a new method for parallel incremental decomposition designed with two main goals in mind: \textbf{G1}:  Compression, speed , simplicity, and parallelization; and \textbf{G2}: correctness in recovering compressed partial results for incoming data, under suitable conditions. The algorithmic framework we propose is shown in Figure \ref{fig:crown_jewel} and is described below:

We assume that we have a pre-existing set of {\em summaries} of the $\tensor{X}$ before the update. Summaries are in the form of compressed tensors of dimension [$Q \times Q \times Q$].

These are generated by multiplying random compression matrices \{$\mathbf{U}, \mathbf{V}, \mathbf{W}$\} that are independently obtain from an absolutely continuous uniform distribution with respect to the Lebesgue measure, with tensor's corresponding mode i.e. $\mathbf{U}$ is multiplied with $I$-mode and so on; see Figure \ref{fig:crown_jewel} and Section \ref{subsec:factormatch} for its role in correct identification of factors. The compression matrices are generated for each incoming batch or slice. For simplicity of description, we assume that we are receiving updated slice(s) on the third mode. We, further, assume that the updates come in batches of new slices, which, in turn, ensures that we see a mature-enough update to the tensor, which contains useful structure. Trivially, however, \method can operate on singleton batches and for $>3$ modes also.

In the following, $\tensor{X}_{old}$ is the tensor prior to the update and $\tensor{X}_{new}$ is the batch of incoming slice(s). Considering $S=\prod_{i=1}^{[N-1]}I^{(i)}$ and $T=\sum_{i=1}^{[N-1]}I^{(i)}$ , we can write space and time complexity in terms of $S$ and $T$. Given an incoming batch, \method performs the following steps:

\subsection{\textbf{Parallel Compression and Decomposition}} 
When handling large tensors $\tensor{X}$ that are unable to fit in main memory, we may compress the tensor $\tensor{X}$ to a smaller tensor that somehow apprehends most of the systematic variation in  $\tensor{X}$. Keeping this in mind, for incoming slice(s) $\tensor{X}_{new}$, during the parallel compression step, we first need to create '$p$' parallel triplets of random compression matrices (uniformly distributed) \{$\mathbf{U}_p, \mathbf{V}_p, \mathbf{W}_p$\} of $\tensor{X}$.  Thus, each worker (i.e. Matlab parpool) is responsible for creating and storing these triplets of size $\mathbf{U} \in \mathbb{R}^{I \times Q},\ \ \mathbf{V}\in \mathbb{R}^{J \times Q}$ and $\mathbf{W}\in \mathbb{R}^{t_{new} \times Q}$. These matrices share at least 'shared' amount of column(s) among each other. Mathematically, we can describe it as follows:
\begin{equation}
\small
\tensor{X}=  \begin{bmatrix}
\{\mathbf{U}_1, \mathbf{V}_1,\mathbf{W}_1\} \\
\{\mathbf{U}_2, \mathbf{V}_2, \mathbf{W}_2\}\\
\dots\\
\{\mathbf{U}_p, \mathbf{V}_p, \mathbf{W}_p\}
\end{bmatrix} 
= \begin{bmatrix}
\{(\mathbf{u} \ \ \mathbf{U}_{1'}),(\mathbf{v} \ \ \mathbf{V}_{1'}),(\mathbf{w}  \ \ \mathbf{W}_{1'})\} \\
\{(\mathbf{u} \ \ \mathbf{U_{2'}}),(\mathbf{v} \ \ \mathbf{V}_{2'}),(\mathbf{w}  \ \ \mathbf{W}_{2'})\} \\
\dots\\
\{(\mathbf{u} \ \ \mathbf{U}_{p'}),(\mathbf{v} \ \ \mathbf{V}_{p'}),(\mathbf{w}  \ \ \mathbf{W}_{p'})\} \\
\end{bmatrix}
\end{equation}
where u, v and w are shared and have dimensions of $\mathbb{R}^{I \times Q_{shared}}, \mathbb{R}^{J \times Q_{shared}}$ and $\mathbb{R}^{t_{new} \times Q_{shared}}$.

For compression matrices, we choose to assign each worker create a single row of each of the matrices to reduce the burden of creating an entire batch of \{$\mathbf{U}_p^{'}, \mathbf{V}_p^{'}, \mathbf{W}_p^{'}$\} of $\tensor{X}_{new}$. We see that each worker is sufficient to hold \hide{\{$\mathbf{U}_p, \mathbf{V}_p, \mathbf{W}_p$\}} these matrices in main memory. Now, we created compressed tensor replica or summaries $\{\tensor{Z}_1, \tensor{Z}_2 \dots \tensor{Z}_p\}$ by multiplying each triplets of compression matrices and $\tensor{X}_{new}$;see Figure \ref{fig:crown_jewel}.  $\tensor{Z}_{p}$ is 3-mode tensor of size  $\mathbb{R}^{Q \times Q \times Q}$. Since $Q$ is considerably smaller than [I ,J,  K], we use $ O(Q^3)$ of memory on each worker. 

For $\tensor{X}_{old}$, we already have replicas $\{\tensor{Y}_1, \tensor{Y}_2 \dots \tensor{Y}_p\}$ obtained from each triplets of compression matrices \{$\mathbf{U}_p, \mathbf{V}_p, \mathbf{W}_p$\}and $\tensor{X}_{old}$;see Figure \ref{fig:crown_jewel}. In general, the compression comprises N-mode products which leads to overall complexity of $(Q_{(1)}St_{new} + Q_{(2)}St_{new} +Q_{(3)}St_{new} + \dots Q_{(N)}St_{new})$ for dense tensor $\tensor{X}$, if the first mode is compressed first, followed by the second, and then the third mode and so on. We choose to keep $Q_1, Q_2,Q_3 \dots Q_N$ of same order as well non-temporal dimensions are of same order in our algorithm, so time complexity of parallel compression step for N-mode data is $O(QSt_{new})$ for each worker. The {\em summaries} are always dense, because first mode product with tensor is dense, hence remaining mode products are unable to exploit sparsity. 

After appropriately computing {\em summaries} $\{\tensor{Z}_1, \tensor{Z}_2 \dots \tensor{Z}_p\}$ for incoming slices, we need to update the old summaries $\{\tensor{Y}_1, \tensor{Y}_2 \dots \tensor{Y}_p\}$ which were generated from previous data. We don't save entire $\tensor{X}_{old}$, and instead we save the compressed summaries i.e. $\tensor{Y}$ only. Each worker reads its segment and process update in parallel as given below. 
   
\begin{equation}
\ssmall
     \begin{bmatrix}
    \tensor{X}_1 \\
    \tensor{X}_2\\
    \vdots\\
    \tensor{X}_p
    \end{bmatrix} 
    = 
     \begin{bmatrix}
    \tensor{Y}_1 \\
    \tensor{Y}_2\\
    \vdots\\
   \tensor{Y}_p
    \end{bmatrix}
   \oplus 
    \begin{bmatrix}
    \tensor{Z}_1 \\
    \tensor{Z}_2\\
    \vdots\\
    \tensor{Z}_p
    \end{bmatrix} \odot  \begin{bmatrix}
    \mathbf{W}^{'}_{1}(k,q) \\
    \mathbf{W}^{'}_{2}(k,q)\\
    \vdots\\
    \mathbf{W}^{'}_{p}(k,q)
    \end{bmatrix} = \begin{bmatrix}
    (\mathbf{A}_{s(1)}, \mathbf{B}_{s(1)} , \mathbf{C}_{s(1)}) \\
    (\mathbf{A}_{s(2)}, \mathbf{B}_{s(2)} , \mathbf{C}_{s(2)})\\
    \vdots\\
    (\mathbf{A}_{s(p)}, \mathbf{B}_{s(p)} , \mathbf{C}_{s(p)})
    \end{bmatrix}
\end{equation}            
 where $k$ is the number of slices of incoming tensor and $q$ is the slice number for the compressed tensor. Further, for the decomposition step, we processed '$p$' {\em summaries} on different workers, each one fitting the decomposition to the respective compressed tensor $\{\tensor{X}_1, \tensor{X}_2 \dots \tensor{X}_p\}$ created by the compression step.  We assume that the updated compressed tensor $\{\tensor{X}_1, \tensor{X}_2 \dots \tensor{X}_p\}$ fits in the main memory, and performs in-memory computation.  We denote $p^{th}$ compressed tensor decompositions as $(\mathbf{A}_{s(p)}, \mathbf{B}_{s(p)} , \mathbf{C}_{s(p)})$ as discussed above. The data for each parallel worker $\tensor{X}_p$ can be uniquely decomposed, i.e. $(\mathbf{A}_p,\mathbf{B}_p,\mathbf{C}_p)$ is unique up to scaling and column permutation.
Furthermore, parallel compression and decomposition is able to achieve Goal \textbf{G1}.

\subsection{\textbf{Factor match for identifiability}}
\label{subsec:factormatch}
 According to Kruskal \cite{harshman1972determination}, the CP decomposition is unique (under mild conditions) up to permutation and scaling of the components i.e. $\mathbf{A},\mathbf{B}$ and $\mathbf{C}$ factor matrices. Consider an 3-mode tensor $\tensor{X}$ of dimension $I$, $J$ and $K$ of rank $R$. If rank 
\begin{equation}
\label{equ:uniq1}
r_c=F \implies K \geq R \  \&  \ I(I-1)(J-1)\geq 2R(R-1),
\end{equation}
 then rank 1 factors of tensor $\tensor{X}$  can be uniquely computable\cite{harshman1972determination,jiang2004kruskal}. \hide{With help of }Kronecker product\cite{brewer1978kronecker} property is described as  $(\mathbf{U}^T \otimes \mathbf{C}^T \otimes \mathbf{W}^T)(\mathbf{A} \odot\mathbf{B} \odot \mathbf{C}) = ((\mathbf{U}^T\mathbf{A})\odot(\mathbf{V}^T\mathbf{B})\odot(\mathbf{W}^T\mathbf{C})) \approx (\widetilde{\mathbf{A}} ,\widetilde{\mathbf{B}} ,\widetilde{\mathbf{C}})$. Now combining Kruskal's uniqueness and Kronecker product property, we can obtain correct identifiable factors from {\em summaries} if 
\begin{equation} 
\label{equ:uniq2}
 \min(Q,r_A)+\min(Q,r_B)+\min(Q,r_C) \geq 2R+2
\end{equation}
 where Kruskal-rank of $\mathbf{A}$, denoted as $r_A$, is the maximum $r$ such that any $r$ columns of $\mathbf{A}$  are linearly independent;see \cite{sidiropoulos2014parallel}. Hence, upon factorization of 3-mode $\tensor{X_p}$ into $R$ components, we obtain $ \mathbf{A}= a^T_pA\Pi_{p}\lambda_p^{(1/N)}$ where $a$ is shared among summaries decompositions , $\Pi_p$ is a permutation matrix, and $\lambda_p$ is a diagonal scaling matrix obtained from CP decomposition. To match factor matrices after decomposition step, we first normalize the shared columns of factor matrices $(\mathbf{A}_{s(i)}, \mathbf{B}_{s(i)} , \mathbf{C}_{s(i)})$ and $(\mathbf{A}_{s(i+1)}, \mathbf{B}_{s(i+1)} , \mathbf{C}_{s(i+1)})$ to unit norm $||.||_1$ \hide{From V: I think you need $\| \|_2$ here, right?, From Ekta: here we need to normalize each column by dividing it with its max value. Basically converting values between 0-1.}. Next, for each column of $(\mathbf{A}_{s(i+1)}, \mathbf{B}_{s(i+1)}, \mathbf{C}_{s(i+1)})$, we find the most similar column of $(\mathbf{A}_{s(i)},  \mathbf{B}_{s(i)} , \mathbf{C}_{s(i)})$, and store the correspondence. Finally, we can describe factor matrices as :
 
\begin{equation}
\ssmall
 \mathbf{\widetilde{A}_s} = \begin{bmatrix}
        a^T_1\\
        a^T_2\\
 	\vdots\\
 	a^T_p
 \end{bmatrix} * \mathbf{\widetilde{A}} \Pi\lambda^{(1/N)} , \  \ \mathbf{\widetilde{B}_s} = \begin{bmatrix}
 b^T_1\\
 b^T_2\\
 \vdots\\
b^T_p
 \end{bmatrix} * \mathbf{\widetilde{B}} \Pi\lambda^{(1/N)} , \  \ \mathbf{\widetilde{C}_s} = \begin{bmatrix}
 c^T_1\\
 c^T_2\\
 \vdots\\
 c^T_p
 \end{bmatrix} * \mathbf{\widetilde{C}} \Pi\lambda^{(1/N)} 
\end{equation}
 
where $\widetilde{\mathbf{A}}_s,\widetilde{\mathbf{B}}_s,$ and $\widetilde{\mathbf{C}}_s$ are matrices of dimension $\widetilde{\mathbf{A}}_s \in \mathbb{R}^{pQ \times R}, \ \ \widetilde{\mathbf{B}}_s \in\mathbb{R}^{pQ \times R}$ and $\widetilde{\mathbf{C}}_s \in \mathbb{R}^{pQ \times R}$ respectively and N is number of dimensions of tensor. For 3-mode tensor, $N = 3$ and for 4-mode tensor, $N = 4$ and so on. Even though for 3-mode tensor, $\mathbf{A}$ and $\mathbf{B}$ do not increase their number of rows over time, the incoming slices may contribute valuable new estimates to the already estimated factors.Thus, we update all factor matrices in the same way. This is able to partially achieve Goal \textbf{G2}.

\subsection{\textbf{Update results}}  Final step is to remove all the singleton dimensions from the sets of compression matrices \{$\mathbf{U}_p,\mathbf{V}_p,\mathbf{W}_p$\} and stack them together. A singleton dimension of tensor or matrix is any dimension for which size of matrix or tensor with given dimensions becomes one. Consider the 5-by-1-by-5 array A. After removing its singleton dimension, the array A become 5-by-5. Now, we can write compression matrices as:
\begin{equation}
\ssmall
	\widetilde{\mathbf{P}}_a= \begin{bmatrix}
	\mathbf{U(:,:,1)^T}\\
   \mathbf{	U(:,:,2)^T}\\
	\vdots\\
\mathbf{	U(:,:,p)^T}
	\end{bmatrix}  , \  \ \widetilde{\mathbf{P}}_b = \begin{bmatrix}
	\mathbf{V(:,:,1)^T}\\
	\mathbf{V(:,:,2)^T}\\
	\vdots\\
	\mathbf{V(:,:,p)^T}
	\end{bmatrix} , \  \ \widetilde{\mathbf{P}}_c = \begin{bmatrix}
	\mathbf{W(:,:,1)^T}\\
	\mathbf{W(:,:,2)^T}\\
	\vdots\\
\mathbf{	W(:,:,p)^T}
	\end{bmatrix} 
\end{equation}

where $\widetilde{\mathbf{P}}_a,\widetilde{\mathbf{P}}_b,$ and $\widetilde{\mathbf{P}}_c$ are matrices of dimension $\widetilde{\mathbf{P}}_a \in \mathbb{R}^{pQ \times I},\ \ \widetilde{\mathbf{P}}_b \in \mathbb{R}^{pQ \times J}$ and $\widetilde{\mathbf{P}}_c \in \mathbb{R}^{pQ \times K}$ respectively. The updated factor matrices ($\mathbf{A}$,$\mathbf{B}$, and $\mathbf{C}$) for 3-mode tensor $\tensor{X}$ (i.e. $\tensor{X}_{old} +  \tensor{X}_{new}$)  can be obtained by : 

\begin{equation}
\mathbf{A}=\widetilde{\mathbf{P}}_a^{-1}*\widetilde{\mathbf{A}}_s,\ \ 
\mathbf{B}= \widetilde{\mathbf{P}}_b^{-1}*\widetilde{\mathbf{B}}_s , \ \ \mathbf{C}= \big[ \mathbf{C}_{old};\widetilde{\mathbf{P}}_c^{-1}*\widetilde{\mathbf{C}}_s\big]
\end{equation}

where $\mathbf{A},\mathbf{B}$ and $\mathbf{C}$ are matrices of dimension $\mathbb{R}^{I \times R}, \ \ \mathbb{R}^{J \times R}$ and $\mathbb{R}^{K_{1\dots n,(n+1) \dots m} \times R}$ respectively. Hence, we achieve Goal \textbf{G2}.

%	\vspace{-0.5in}

Finally, by putting everything together, we obtain the general version of our \method for 3-mode tensor, as presented in Algorithm \ref{alg:method} in supplementary material. The matlab implementation for method is available at $link^{1}$. The \textbf{\em{higher order}} version of \method is also given in supplementary materials. We refer the interested reader to supplementary material that provide more details on \method and its applications.

\textbf{Complexity Analysis}: As discussed previously, compression step's time and space complexity is $O(QSt_{new})$ and $O(Q^3)$ respectively.  Identifiability and update can be calculated in $O(pQI+pQR)$. Hence, time complexity is considered as $O(p^2QI+p^2QIR+QSt_{new})$. Overall, as S is larger than any other factors, the time complexity of \method can be written as $O(QSt_{new})$. In terms of space consumption, \method is quite efficient since only the compressed matrices, previous factor matrices and summaries need to be stored. Hence, the total cost of space is $pQ(pT+t_{new}+R)+(T+t_{old})R+Q^3$.
\hide{
\begin{table*}[h!]
	\ssmall
	\begin{center}
		\begin{tabular}{ |c||c|c|c| }
			\hline
			Methods & Time Complexity & Space Complexity & Reference \\
			\hline
			\method&$O(QSt_{new})$&$pQ(pT+t_{new}+R)+(T+t_{old})R+Q^3$&\\
			OnlineCP&$O(NSt_{new}R)$&$St_{new}+(2T+t_{old})R+(N-1)R^2$&\cite{zhou2016accelerating}\\
			SambaTen&$O(nnz(X)R(N-1)$& $nnz(X)+(T+t_{old}+2S)R+2R^2$&\cite{gujral2017sambaten}\\
			RLST&$O(R^2S)$&$St_{new}+(T+t_{old}+2S)R+2R^2$&\cite{nion2009adaptive}\\
			ParaComp&$O(QS(t_{old}+t_{new}))$&$St_{new}+(T+t_{old})R+pQ^3$&\cite{sidiropoulos2014parallel}\\
			\hline
		\end{tabular}
		\caption{\textbf{Complexity comparison between \method and state-of-art methods.}}
		\label{table:complexAnalysis}
	\end{center}
	\vspace{-0.2in}
\end{table*}

}
\hide{
\subsection{\textbf{Necessary characteristics for uniqueness}}
As we mention above, \method is able to identify the solution of the online CP decomposition, as long as the parallel CP decompositions on the compressed tensors are also identifiable. Empirically, we observe that if the decomposition\hide{was} of a given data that has \hide{have} exact or near-trilinear structure (or multilinear in the general case), i.e. obeying the low-rank CP model with some additive noise, \method is able to successfully, accurately, and using much fewer resources than state-of-the-art, track the online decomposition. On the other hand, when given data that do not have a low trilinear rank, the results were of lower quality. This observation is of potential interest in exploratory analysis, where we do not know 1) the (low) rank of the data, and 2) whether the data have low rank to begin with (we note that this is an extremely hard problem, out of the scope of this paper, but we refer the interested reader to previous studies \cite{wang2018low,papalexakis2016automatic} for an overview). If \method provides a good approximation, this indirectly signifies that the data have appropriate trilinear structure, thus CP is a fitting tool for analysis. If, however, the results are poor, this may indicate that we need to reconsider the particular rank used, or even analyzing the data using CP in the first place. We reserve further investigations of what this observation implies for future work.
}
%It is important to eliminate the unnecessary constants for two-way interactions before fitting any tri-linear model. Failure to do so, results in serious interfere with the identification of parallel components after decomposition. This problem becomes more profound when 3-way data are considered. It is not necessary that all possible transformations are appropriate. Finding the transformation of data that does not disturb the tri-linear structure of the data has been a subject of considerable study. When the data are more or less tri-linear then the decompositions are generally better and method converges very fast and are unique and interpretable or identifiable. If after decomposition step, the latent factors do not show distinct variation in its respective modes, then it will fail to have a uniquely identified solution. Hence, uniqueness of decomposition depends on permutation of latent components, scaling of the factors and equation \ref{equ:uniq1} - \ref{equ:uniq2}. If these conditions satisfy for the given data, \method is able to find incremental decomposition accurately, fast and consumes less memory when compared to state-of-art techniques.

%% file: 040experiments.tex
\section{Empirical Analysis}
\label{sec:experiments}
We design experiments to answer the following questions: \textbf{(Q1)} How  much memory \method  required for updating incoming data?
\textbf{(Q2)} How fast and accurately are updates in \method compared to incremental algorithms?
\textbf{(Q3)} How does the running time of \method increase as tensor data grow (in time mode)?
\textbf{(Q4)} What is the influence of parameters on \method ?
\textbf{(Q5)} How \method used in real-world scenarios? 

For our all experiments, we used Intel(R) Xeon(R), CPU E5-2680 v3 @ 2.50GHz machine with 48 CPU cores and 378GB RAM. 
\hide{In this section we extensively evaluate the performance of \method on multiple synthetic datasets, and compare its performance with state-of-the-art approaches.We experiment on the different parameters of \method, and how that affects performance } 

% % %***************************Evaluation Measures***************************
\subsection{\textbf{Evaluation Measures}}
\label{sec:EvaMeas}
%In order to obtain an accurate picture of the performance, 
We evaluate \method and the baselines using three criteria: fitness, processor memory used, and wall-clock time. These measures provide a quantitative way to compare the performance of our method. More specifically, \textbf{Fitness} measures the effectiveness of approximated tensor and defined as : 
$$
Fitness(\%)= 100* \Big(1- \frac{||\tensor{X}-\widetilde{\tensor{X}}||_F}{||\tensor{X} ||_F}\Big)
$$
higher the value, better the approximation. Here $\tensor{X}$ is original tensor and $\widetilde{\tensor{X}}$ is drawn tensor from \method.

 \textbf{CPU time (sec)}: indicates the average running time for processing all slices for given tensor, measured in seconds, is used to validate the time efficiency of an algorithm.

\textbf{Process Memory used (MBytes)}: indicates the average memory required to process each slices for given tensor, is used to validate the space efficiency of an algorithm.
% % %***************************Experimental Results***************************

\subsection{\textbf{Baselines}}
In this experiment, four baselines have been selected as the competitors to evaluate the performance. 
\textbf{OnlineCP}\cite{zhou2016accelerating}: It is online CP decomposition method, where the latent factors are updated when there are new data.
\textbf{SambaTen}\cite{gujral2017sambaten}: Sampling-based Batch Incremental Tensor Decomposition algorithm is the most recent and state-of-the-art method in online computation of canonical parafac and perform all computations in the reduced summary space.
\textbf{RLST}\cite{nion2009adaptive}: Recursive Least Squares Tracking (RLST) is another online approach in which recursive updates are computed to minimize the Mean Squared Error (MSE) on incoming slice.
\textbf{ParaComp}\cite{sidiropoulos2014parallel}: an implementation of non-incremental parallel compression based tensor decomposition method. The model is based on parallel processing of randomly compressed and reduced size replicas of the data. Here, we simply re-compute decomposition after every update.
%%%%%%%%%%%%%%%%%%%%%%%%%%%%%%%%%%%%%%%% space for tables%%%%%%%%%%%%%%%
% % %***************************Data-set description***************************
\subsection{\textbf{Experimental Setup}}
\label{sec:syntDataDes}
The specifications of each synthetic dataset are given in Table \ref{table:tsyndataset}.
We generate tensors of dimension I = J = K with increasing I and other modes, and added gaussian distributed noise. Those tensors are created from a known set of randomly (uniformly distributed) generated factors with known rank $R$, so that we have full control over the ground truth of the full decomposition. For real datasets , we use AUTOTEN \cite{papalexakis2016automatic} to find rank of tensor. We dynamically calculate the size of incoming batch or incoming slice(s) for our all experiments to fit the data into 10\% of memory of machine. Remaining machine CPU memory is used for computations for algorithm. We use 20 parallel workers for every experiment.
\begin{table}[h!]
	\ssmall
	\begin{center}
		\caption{Table of Datasets analyzed}
		\label{table:tsyndataset}
		\begin{tabular}{ |c||c|c|c|c|c|c|c| }
			\hline
			I=J=K & NNZ&  Batch size  &p&Q&shared&Noise ($\mu,\sigma$) \\
			\hline
			\hline
			50   &$125K$&5 &20&30&5&(0.1, 0.2)\\
			100   &$1M$&10 &30&35&10&(0.2, 0.2)\\
			500  &$125M$&50&40&30&6&(0.5, 0.6)\\
			1000    &$1B$&20&50 &40 &10&(0.4, 0.2) \\
			5000  &$7B$&10&90&70&25&(0.5, 0.6)\\
			10000   &$1T$&10&110&100&20&(0.2, 0.7)\\
			50000  &$6.25T$&4&140&150&30&(0.6, 0.9)\\
				\hline
		\end{tabular}
\end{center}
\vspace{-0.3in}
\end{table}

Note that all comparisons were carried out over 10 iterations each, and each number reported is an average with a standard deviation attached to it. Here, we only care about the relative comparison among baseline algorithms and  it is not mandatory to have the best rank decomposition for every dataset.\hide{Although, there are few methods \cite{morup2009automatic,papalexakis2016automatic} available in literature to find rank of tensor. But for simplicity, we choose the rank R to 5 for all datasets.} In case of  method-specific parameters, for ParaComp algorithm, the settings are chosen to give best performance. For OnlineCP, we choose the batchsize which gives best performance in terms of approximation fitness. For fairness, we also compare against the parameter configuration for \method that yielded the best performance. Also, during processing, for all methods we remove unnecessary variable from baselines to fairly compare with our methods.

%%%%%%%%%%%%%%%%%%%%%%%%%%%%%%%%

\subsection{\textbf{Results}}
\subsubsection{\textbf{[Q1 \& Q2] Memory efficient, Fast and Accurate }}
\label{sec:baslineComp}
\begin{figure*}[!ht]
	
	\begin{center}
		\includegraphics[clip,trim=0.2cm 3.2cm 0cm 2.5cm,width = 0.30\textwidth]{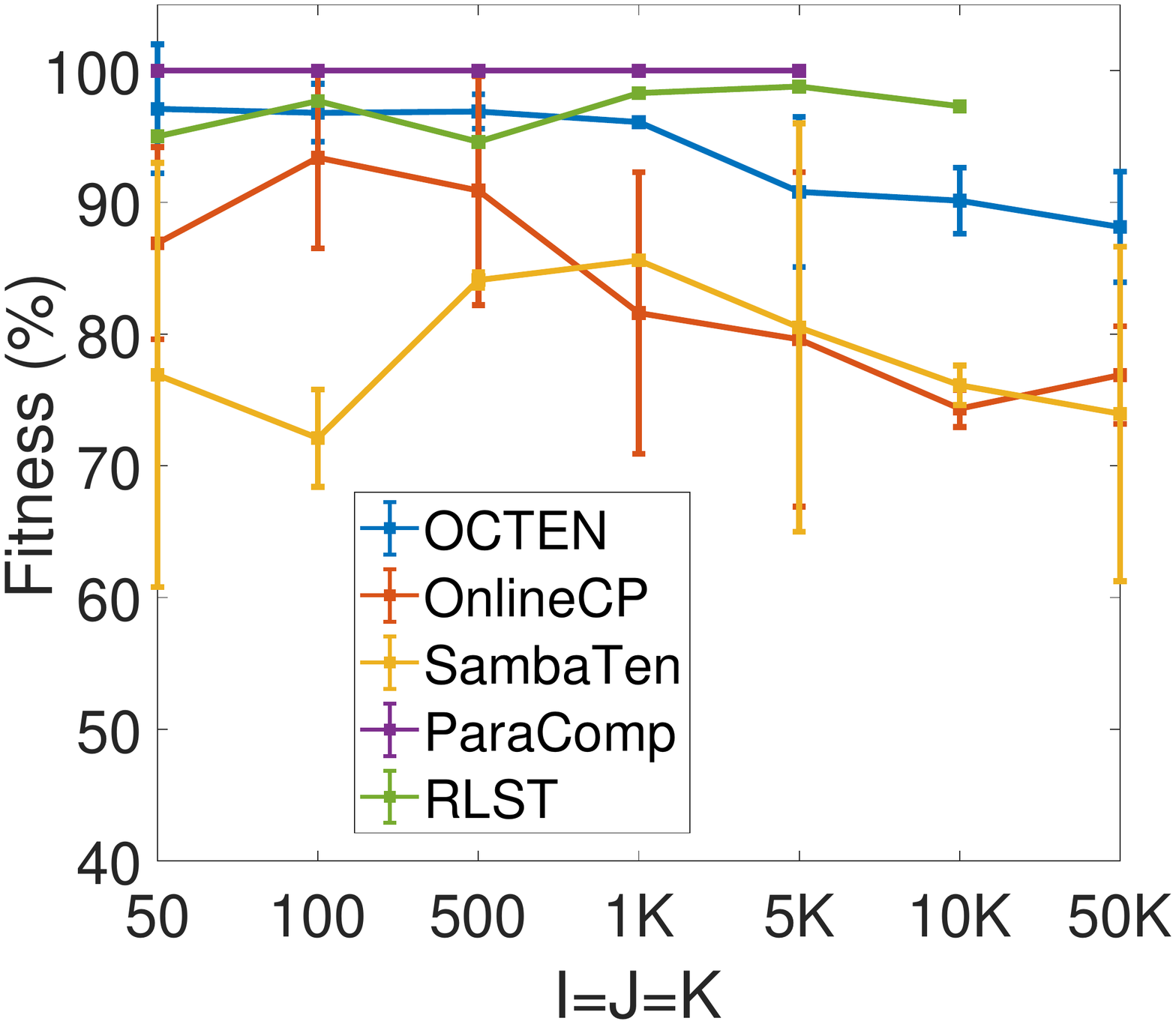}
		\includegraphics[clip,trim=0.2cm 3.2cm 0cm 2.5cm,width = 0.30\textwidth]{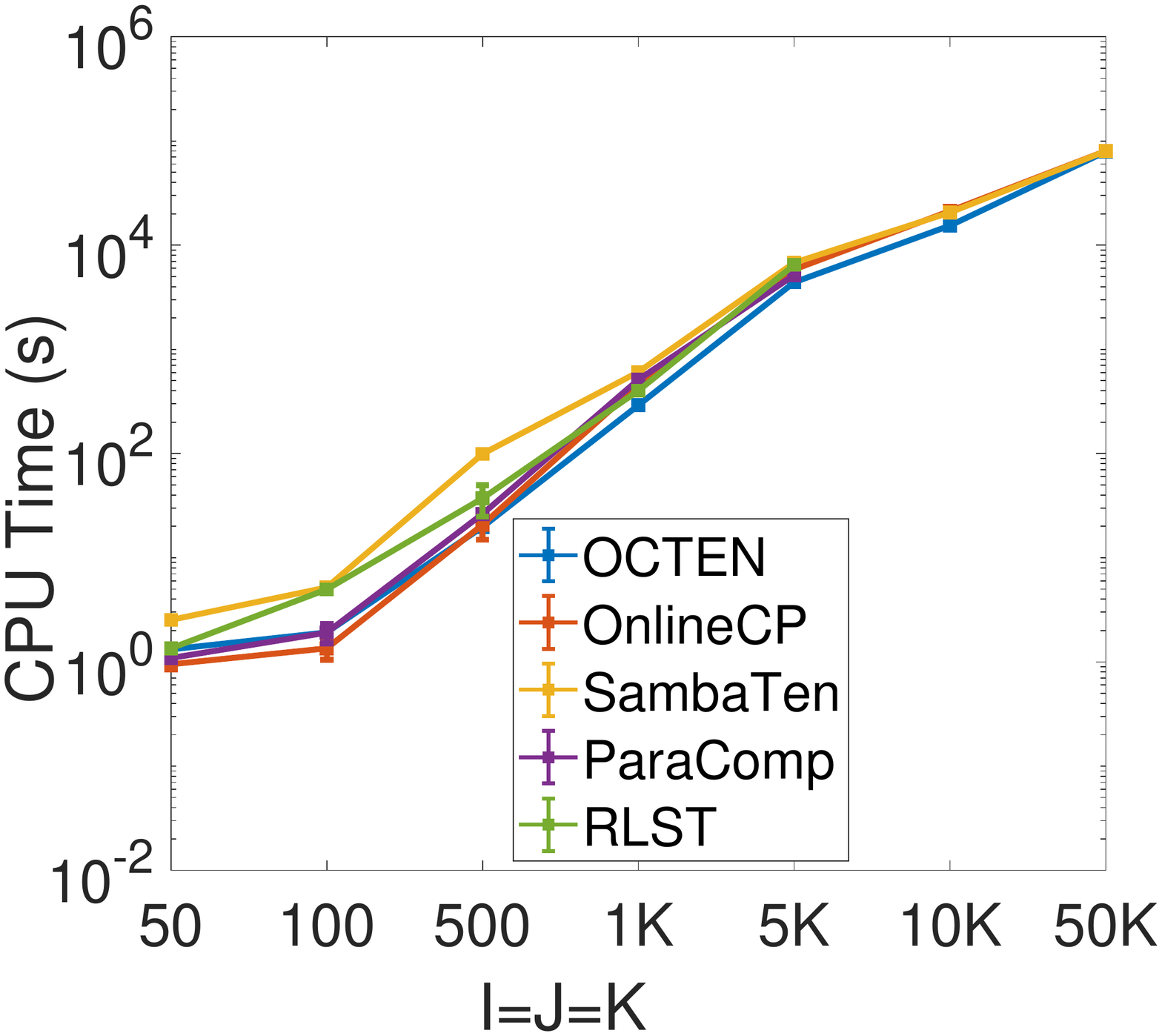}
		\includegraphics[clip,trim=0.2cm 3.2cm 0cm 2.5cm,width = 0.30\textwidth]{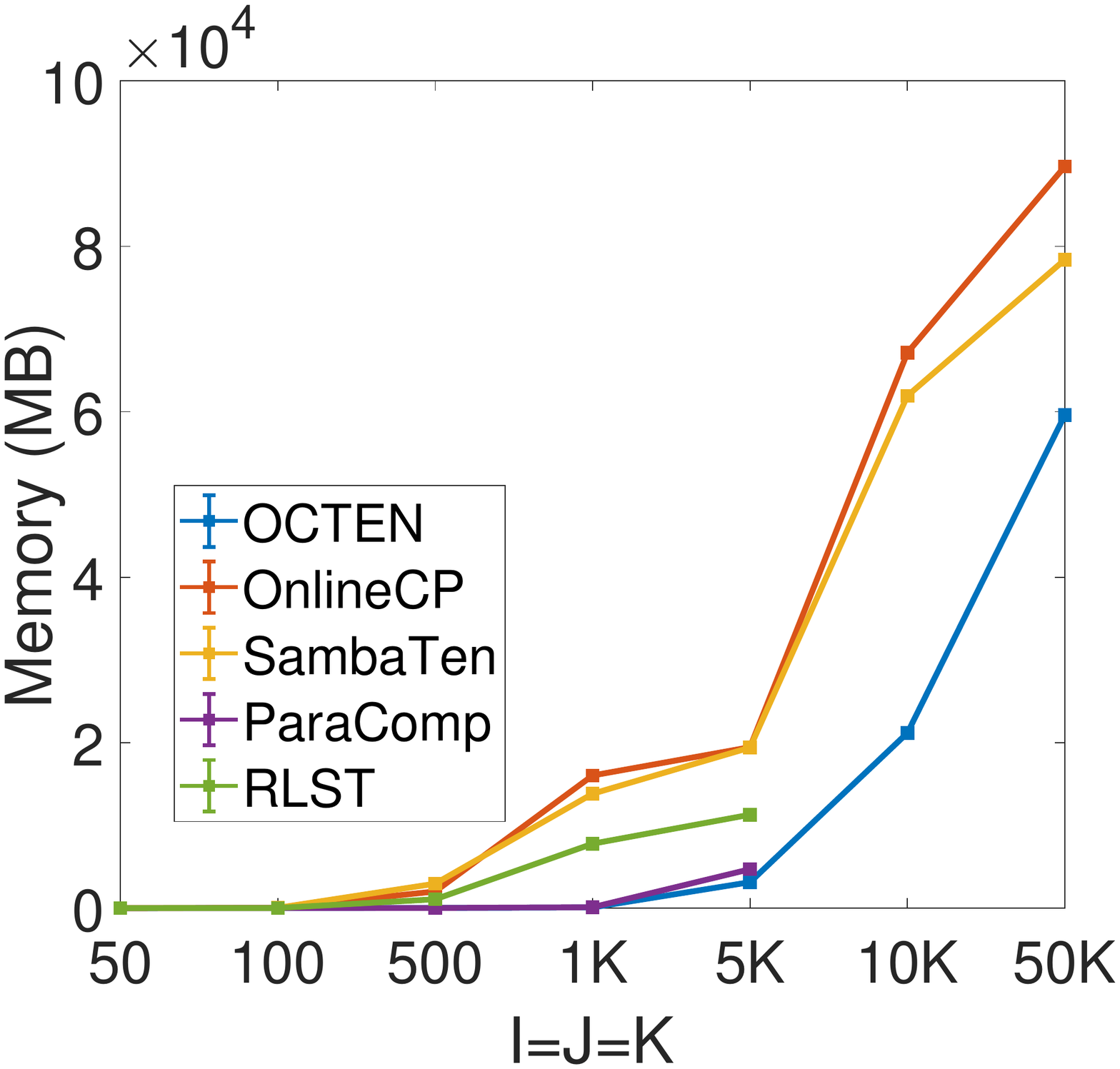}
		\caption{\textbf{(a,b) Experimental results for speed and accuracy of approximation of incoming slices. We see that \method gives comparable accuracy and speed to baseline.(c) Results for memory required to process the incoming slices. The \method remarkably save the memory as compared to baseline methods.This answers our Q1 and Q2.}}
		\label{fig:resultfinal}
	\end{center}
	\vspace{-0.4in}
\end{figure*}

For all datasets we compute Fitness(\%),CPU time (seconds) and Memory(MB) space required. For \method, OnlineCP, ParaComp,Sambaten and RLST we use 10\% of the time-stamp data in each dataset as existing old tensor data. The results for qualitative measure for data is shown in Figure \ref{fig:resultfinal}. For each of tensor data ,the best performance is shown in bold. All state-of-art methods address the issue very well. Compared with OnlineCP, ParaComp,Sambaten and RLST, \method give comparable fitness and reduce the mean CPU running time by up to 2x times for big tensor data. For all datasets, PARACOMP's accuracy (fitness) is better than all methods. But it is able to handle upto $\tensor{X} \in \mathbb{R}^{10^4 \times 10^4 \times 250}$ size only. For small size datasets, OnlineCP's efficiency is better than all methods. For large size dataset, \method outperforms the baseline methods w.r.t fitness, average running time (improved 2x-4x) and memory required to process the updates. It significantly saved 40-200\% of memory as compared to Online CP, Sambaten and RLST as shown in Figure \ref{fig:resultfinal}. It saved 40-80\% memory space compared to Paracomp. Hence, \method is comparable to state-of-art methods for small dataset and outperformed them for large dataset. These results answer Q1 \& Q2 as the \method have comparable qualitative measures to other methods.
\hide{
\begin{table*}[h!]
\ssmall
\setlength{\abovecaptionskip}{15pt plus 3pt minus 2pt}
	\begin{center}
		
		\begin{tabular}{ |c||c|c|c|c|c| }
			\hline
			Dimension &\multicolumn{5}{|c|}{\textbf{Fitness(\%)}} \\
			\hline
		     I=J=K &  \method & OnlineCP& SambaTen&ParaComp&RLST \\
			\hline
			50 &97.1$\pm$4.9&86.9$\pm$7.3&76.9$\pm$16.1&\textbf{100.0$\pm$0} &95.0$\pm$0.1\\
			
			100 &95.8$\pm$2.2&	93.4$\pm$6.9& 72.1$\pm$3.7&	\textbf{100.0$\pm$0} 	&97.7$\pm$0.1\\
			
			500&96.9$\pm$1.3&90.9$\pm$8.7&	84.1$\pm$0.6&\textbf{100.0$\pm$0} &	94.6$\pm$0.1		 \\
			1000&96.1$\pm$0.1&71.6$\pm$10.7&	85.6$\pm$0.1	&\textbf{100.0$\pm$0} &	98.3$\pm$0.1  \\
			5000&90.8$\pm$14.7	&90.6 $\pm$42.7&	80.5$\pm$15.5&	\textbf{100.0$\pm$0} 	&98.8$\pm$0.1		 \\
			10000&\textbf{98.13$\pm$2.5}&54.34$\pm$1.4&56.12$\pm$1.5&NA&97.3$\pm$0.2 \\
			50000&\textbf{68.13$\pm$4.2}&56.89$\pm$3.7&53.95$\pm$12.7&NA&NA \\
	     \hline
		\end{tabular}
	\hide{	\caption{\textbf{Fitness(\%) for incremental tensor decomposition on various datasets. The synthetic datasets have constant rank. Numbers where method outperforms are bolded. Remarkably, We see that \method gives comparable accuracy (Fitness) to baseline.This partially answers our Q2.}}
	\label{table:resultsyndataFitness1}
	\end{center}
\vspace{-0.24in}
\end{table*}

\begin{table*}[h!]
	\ssmall
	\setlength{\abovecaptionskip}{10pt plus 3pt minus 2pt}
	\begin{center}
		}  
	\begin{tabular}{ |c||c|c|c|c|c| }
		\hline
		Dimension &\multicolumn{5}{|c|}{\textbf{CPU Time(sec)}} \\
		\hline
		I=J=K &  \method & OnlineCP& SambaTen&ParaComp&RLST\\
		\hline
		50 &1.32$\pm$0.01&	\textbf{0.95$\pm$0.1}&2.53$\pm$0.1&1.09$\pm$0.1&1.35$\pm$0.1\\	
		100 & 1.92$\pm$0.3&\textbf{1.35$\pm$0.3}&5.19$\pm$0.3&	1.91$\pm$0.3&	4.97$\pm$0.3 \\	
		500	&\textbf{19.74$\pm$0.1}&	20.73$\pm$4.4	&99.21$\pm$2.9	&26.33$\pm$0.1	&37.20$\pm$0.2	\\
		1000&	\textbf{290.91$\pm$6.7}	&455.66$\pm$24.5&	606.26$\pm$1.7&	511.37$\pm$1.6&	401.08$\pm$1.8	\\
		5000&\textbf{4398.89$\pm$45.5}&	5835.46$\pm$0.2K&	6779.71$\pm$0.3K&	5157.98$\pm$3.9&	6464.52$\pm$0.3K\\
		10000&\textbf{15406.55$\pm$156.9}&21287.25$\pm$85.6&20589.81$\pm$69.4&NA&22974.76$\pm$0.1K \\
	    50000&\textbf{78892.80$\pm$98.2}&80159.57$\pm$23.5&79922.80$\pm$47.3&NA&NA \\
	   	\hline
	\end{tabular}  
%	\caption{\textbf{CPU Time (seconds) for incremental tensor decomposition on various datasets. Numbers where method outperforms are bolded. We see that \method gives comparable (similar or better) speed to baseline. However , it is slower when tensor dimensions are small. This partially answers our Q2.}}
\caption{\textbf{Experimental results for speed and accuracy of approximation of incoming slices. We see that \method gives comparable accuracy and speed to baseline.This answers our Q2.}}
		\label{table:resultsyndataFitness}
	\end{center}
\vspace{-0.3in}
\end{table*}
%\vspace{-0.3in}
%%%%%%%%%%%%%%%%%%%%%%%%%% Time CPU %%%%%%%%%%%%%%%%%%%%%%%%%
%%%%% memory %%%%%%%%%%%%%%%%%%%%%%
\begin{table*}[h!]
		\ssmall
		\begin{center}
	 	\begin{tabular}{ |c||c|c|c|c|c| }
			%\hline
			%Dimension & \multicolumn{5}{|c}{\textbf{Process Memory (MB)} }\\
			\hline
			I=J=K & \method & OnlineCP& SambaTen &ParaComp&RLST \\
			\hline
			50  &	\textbf{1.353$\pm$0.001}&	1.509$\pm$0.002&	8.808$\pm$0.001&	5.518$\pm$0.003&	6.271$\pm$0.001\\
			100&	\textbf{13.181$\pm$0.04}	&16.826$\pm$0.03	&30.250	$\pm$0.01&14.525$\pm$0.02	&19.255		$\pm$0.02	 \\
			500&	\textbf{ 25.742$\pm$0.09}&	2018.281$\pm$0.91&	2959.336$\pm$0.15&	26.317$\pm$0.01&	1070.138		$\pm$0.01	 \\
			1000&	\textbf{85.091$\pm$4.1}&	16037.383$\pm$56.5&	13830.850$\pm$21.7&	110.905$\pm$5.1&	7789.841	$\pm$23.2		  \\
			5000 & 	\textbf{3138.052$\pm$10.1}&	19457.678$\pm$25.7	&19409.316$\pm$56.8	&4707.079$\pm$4.1&	11295.270	$\pm$12.9	\\
			10000& \textbf{21179.561$\pm$56.9}&67145.723 $\pm$0.0 &61935.698$\pm$0.0&NA&32459.129	$\pm$456.1  \\
			50000&    \textbf{59613.463$\pm$30.0}   &89657.856 $\pm$0.0 &78387.231$\pm$0.0&NA&NA \\
					\hline
		\end{tabular}
	 \bigskip
	    %\caption{\textbf{Experimental results for memory required to process of incoming slices. Numbers where \method outperforms other baselines are bolded. We see that \method remarkably save the memory as compared to state-of-art techniques.This answers our Q1.}}
	    \caption{\textbf{Experimental results for memory required to process of incoming slices. We see that \method remarkably save the memory as compared to state-of-art techniques.This answers our Q1.}}
	   \label{table:resultsyndataMM}
	\end{center}
\vspace{-0.2in}
\end{table*}
}
%%%%%%%%%%%%%%%%%%%%%%%%%%%%%%%%%%%%%%%% space for tables%%%%%%%%%%%%%%%
%%%%%%%%%%%%%%%%%%%%%%%%%% Scalability Section %%%%%%%%%%%%%%%%%%%%%%%%
\subsubsection{\textbf{[Q3] Scalability Evaluation}}
To evaluate the scalability of our method, firstly, a tensor $\tensor{X} $ of small slice size ($I \in [20,50,100]$) but longer time dimension ($K \in [10^2-10^6]$) is created. Its first $\le$10\% timestamps of data is used for $\tensor{X}_{old}$ and each method's running time for processing batch of $\le$10 data slices at each time stamp measured.

\begin{figure}[!ht]
	\begin{center}
	\includegraphics[clip,trim=0.2cm 5cm 1cm 3cm,width = 0.23\textwidth]{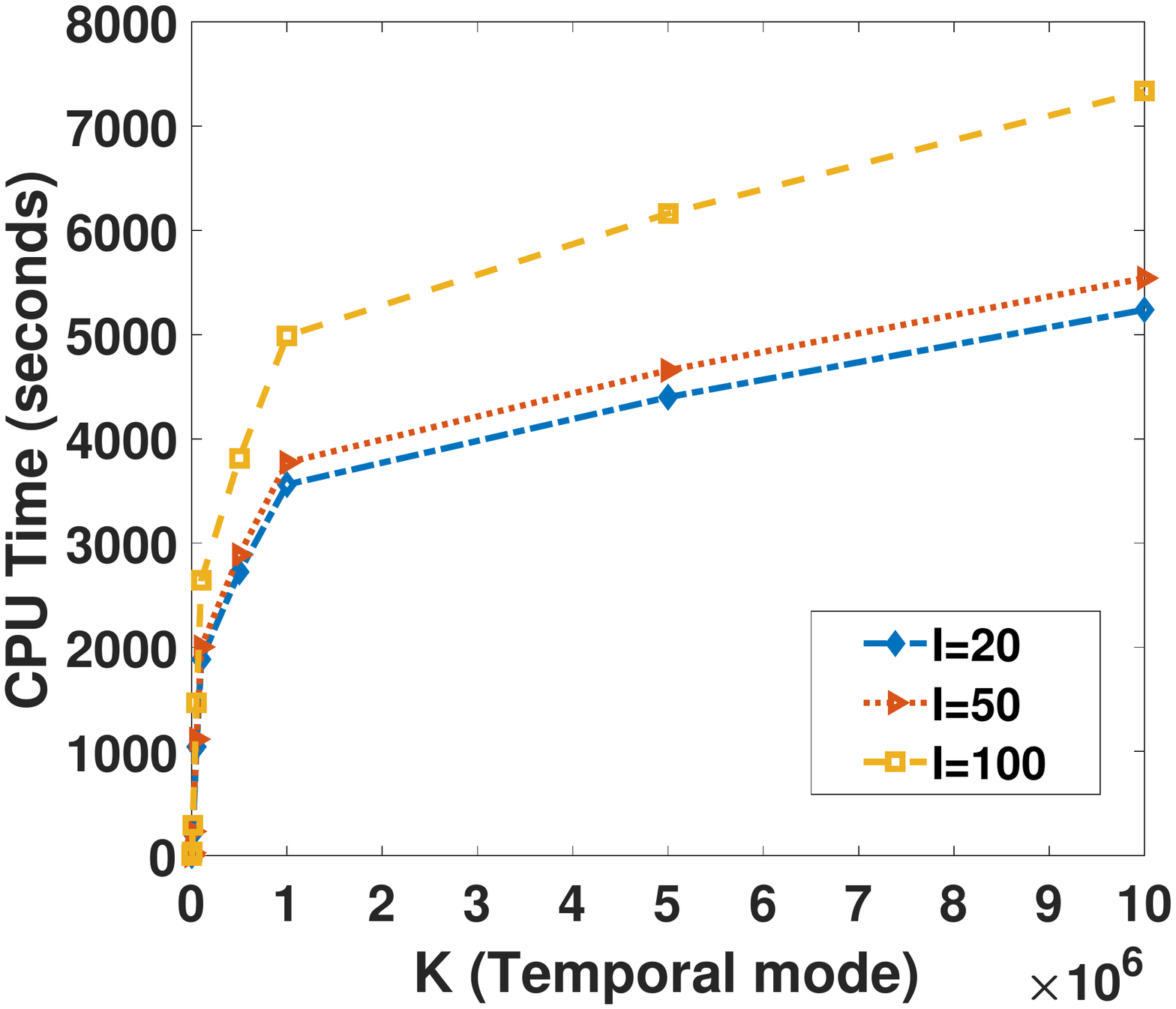}
	\includegraphics[clip,trim=0.2cm 5cm 1cm 3cm,width = 0.23\textwidth]{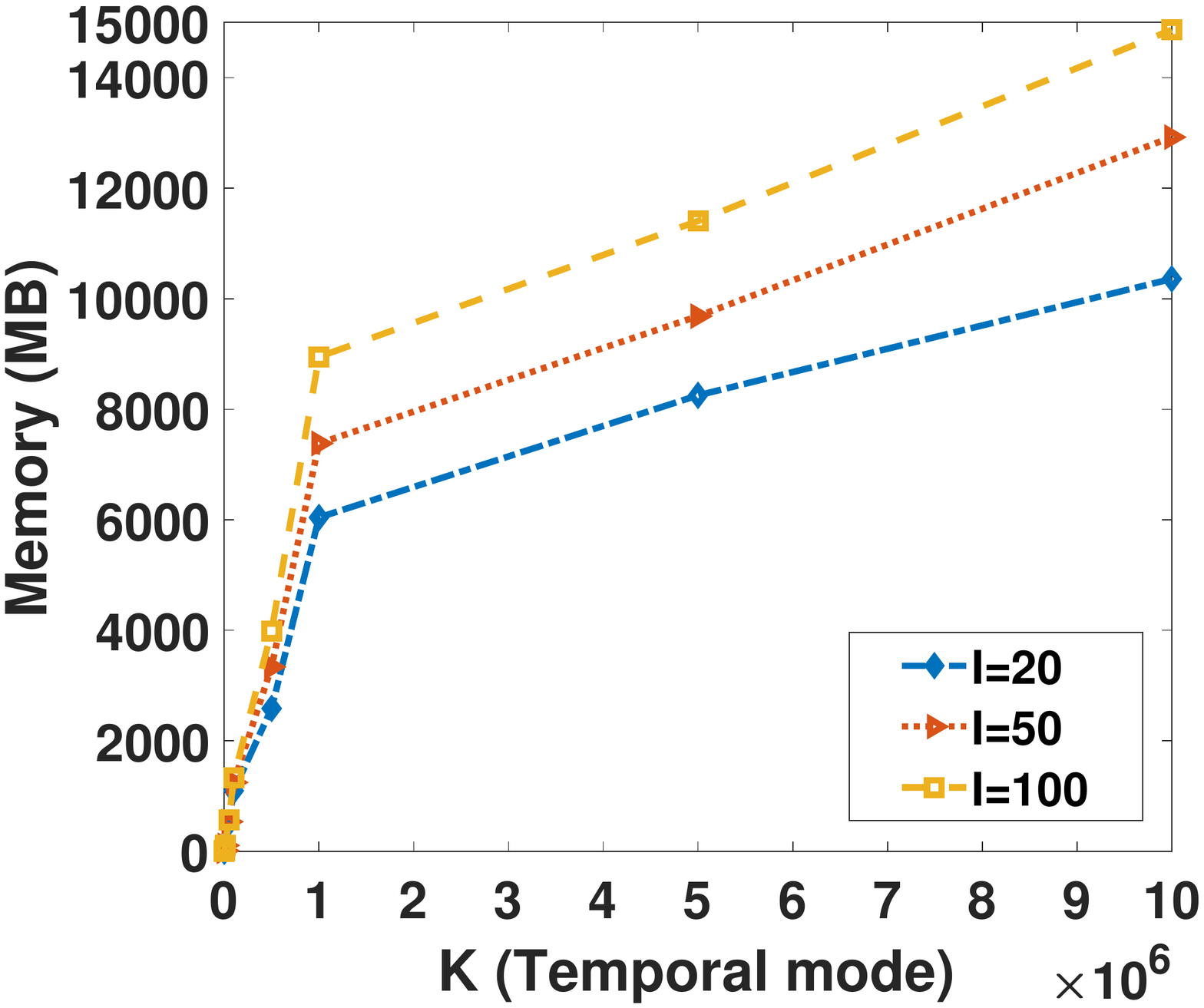}
	\caption{\bf{CPU time (in seconds) and Memory (MB) used for processing slices to tensor $\tensor{X}$ incrementing in its time mode. The time and space consumption increases quasi-linearly. The mean fitness is $\geq$90\% for all experiments}}
	\label{fig:SCALABILITY}
	\end{center}
\vspace{-0.2in}
\end{figure}
As can be seen from Figure \ref{fig:SCALABILITY}, increasing length of temporal mode increases time consumption quasi-linearly. However the slope is different for various non-temporal mode data sizes. In terms of memory consumption, \method also behaves linearly. This is expected behaviour because with increase in temporal mode, the parameters i.e. $p$ and Q also grows. Once again, our proposed method illustrates that it can efficiently process large sized temporal data. This answers our Q3.

 \subsubsection{\textbf{[Q4] Parameter Sensitivity}}
\label{sec:sControl}
We extensively evaluate sensitivity of number of compressed cubes required '$p$' , size of compressed cubes and number of shared columns required for \method. we fixed batch size to $\approx 0.1*K$ for all datasets ,where $K$ is time dimension of tensor . As discussed in section \ref{sec:method}, it is possible to identify unique decomposition . In addition, if we have 
 \begin{equation}
 \label{equ:para}
p \geq \max([\frac{(I-shared)}{(Q-shared)} \ \ \frac{J}{Q}  \ \ \frac{K}{Q}])
\end{equation}
 for parallel workers, decomposition is almost definitely identifiable with column permutation and scaling.
We keep this in mind and evaluate the \method as follows.\\

\begin{figure*}[!ht]
 
	\begin{center}
		\includegraphics[clip,trim=0cm 5cm 0cm 2.5cm,width = 0.25\textwidth]{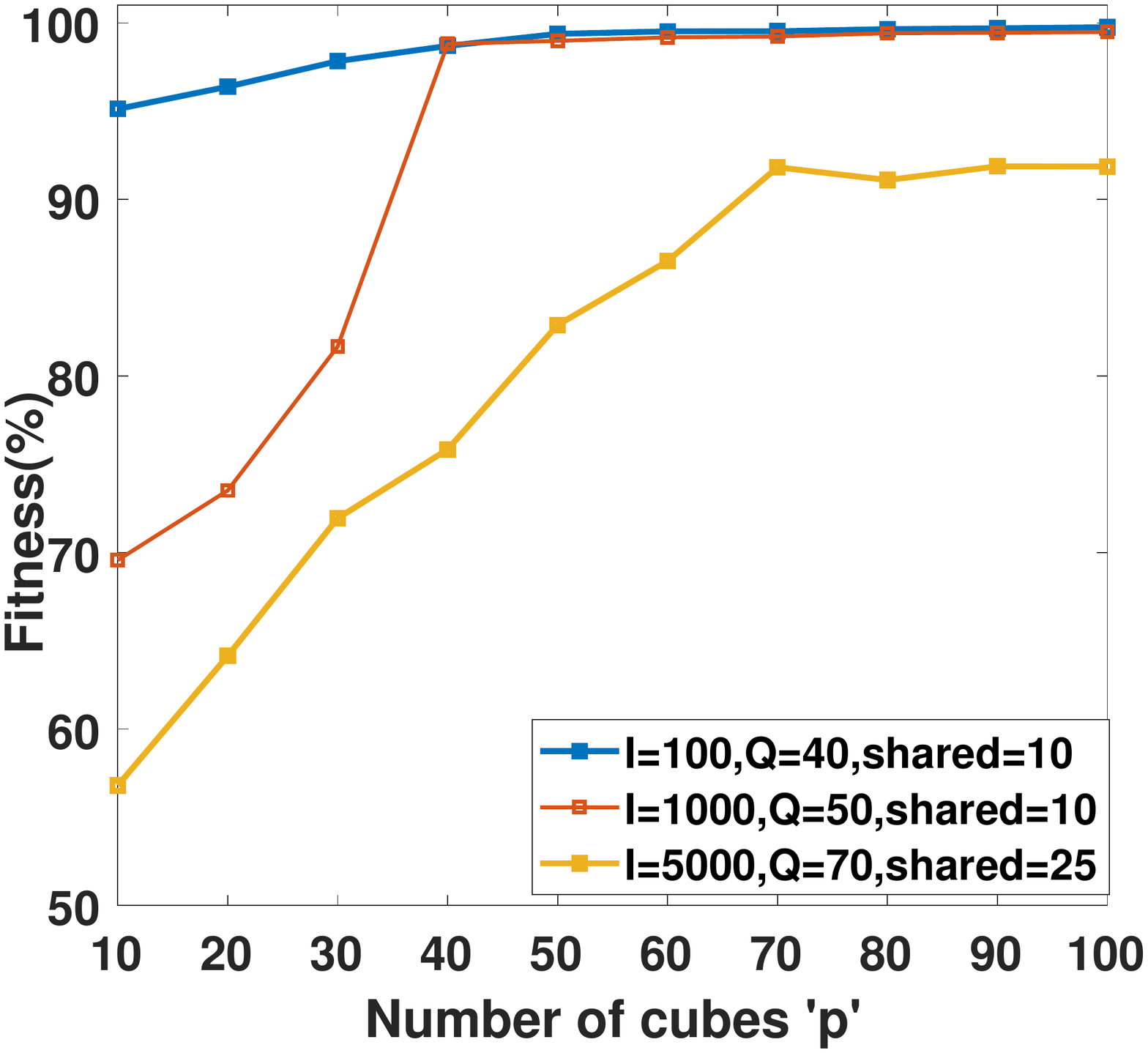}
	    \includegraphics[clip,trim=0cm 5cm 0cm 2.5cm,width = 0.25\textwidth]{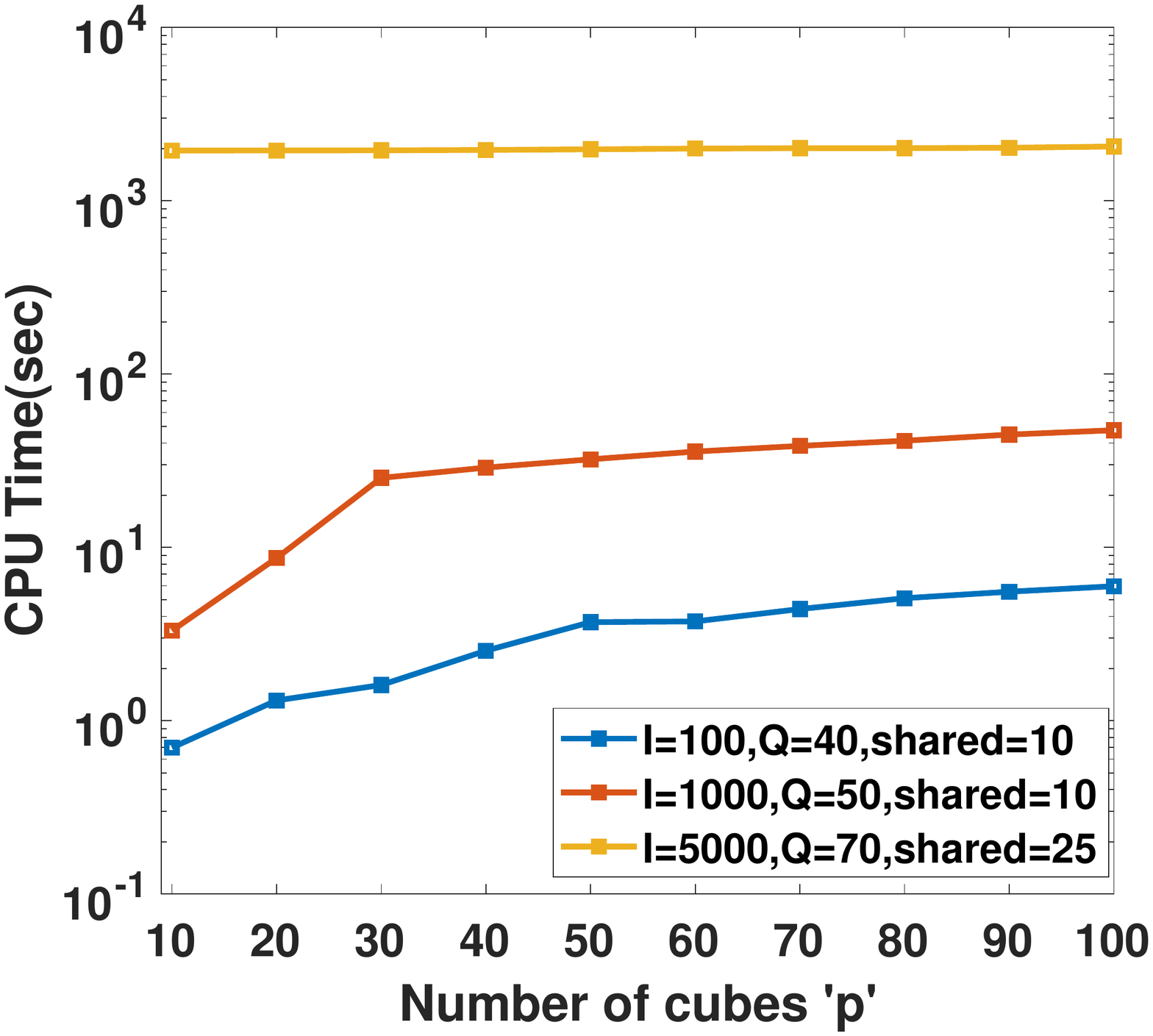}
		\includegraphics[clip,trim=0cm 5cm 0cm 2.5cm,width = 0.25\textwidth]{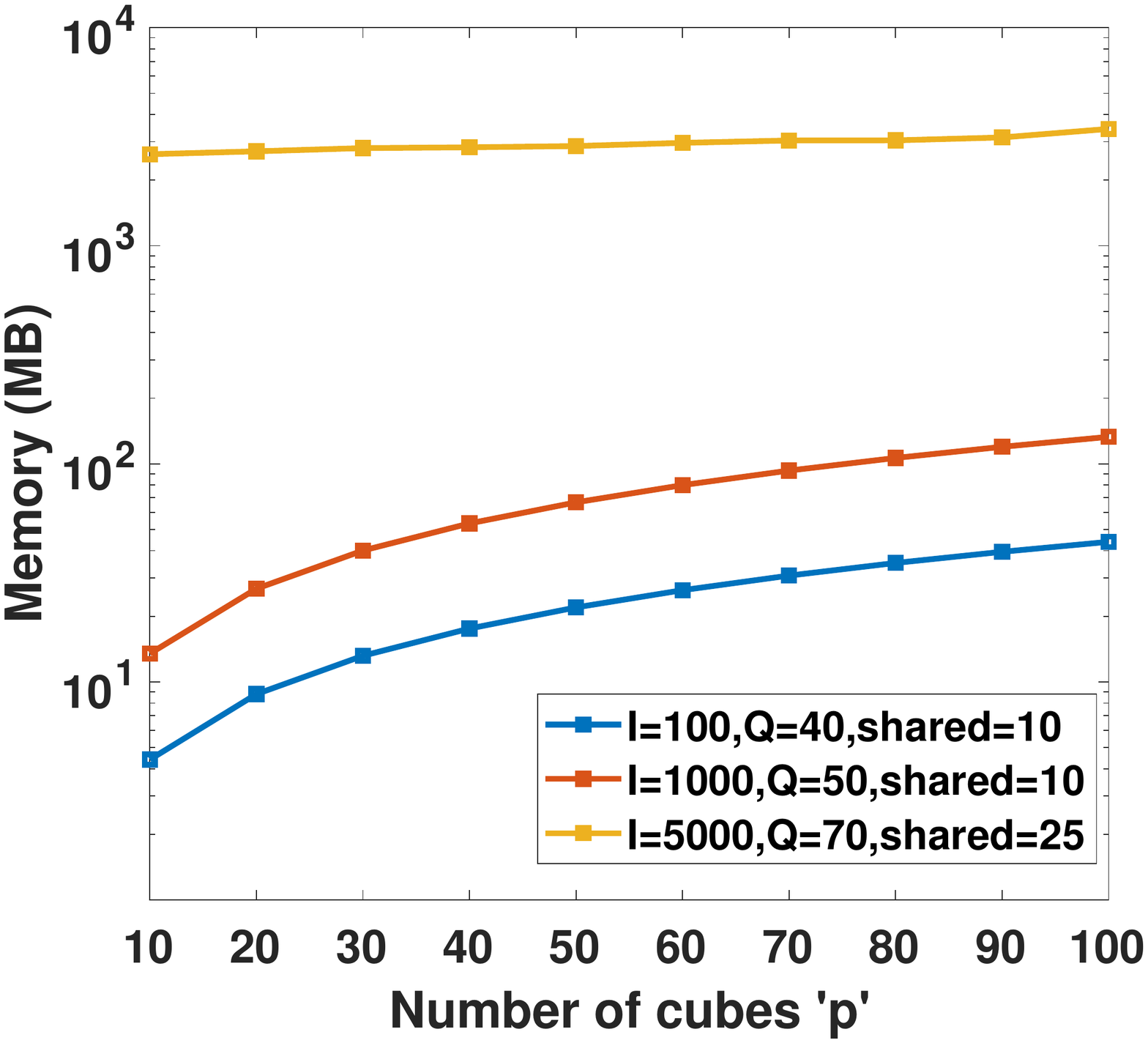}
		\caption{\bf{\method Fitness, CPU Time (sec) and memory used vs. Number of compressed tensors '$p$' on different datasets. With large '$p$', high fitness is achieved.}}
		\label{fig:sen_p}
	\end{center}
 	\vspace{-0.2in}
\end{figure*}

\textbf{(a) Sensitivity of \textit{p}} :The number of compressed cubes play an important role in \method. We performed experiments to evaluate the impact of changing the number of cubes i.e. $p$ with fixed values of other parameters for different size of tensors. We see in figure \ref{fig:sen_p}  that increasing  number of cubes result in increase of Fitness of approximated tensor and  CPU Time and Memory (MB) is super-linearly increased. Consider the case of $I=J=K=1000$, from above condition, we need $P\geq\max\big( \big[ \frac{1000-10}{50-10} \ \ \frac{1000}{50} \ \ \frac{100}{50} \big]\big) \approx 25$. We can see from Figure \ref{fig:sen_p}, the condition holds true.\\

\begin{figure*}[!ht]
 
	\begin{center}
		\includegraphics[clip,trim=0cm 5cm 0cm 2.5cm,width = 0.25\textwidth]{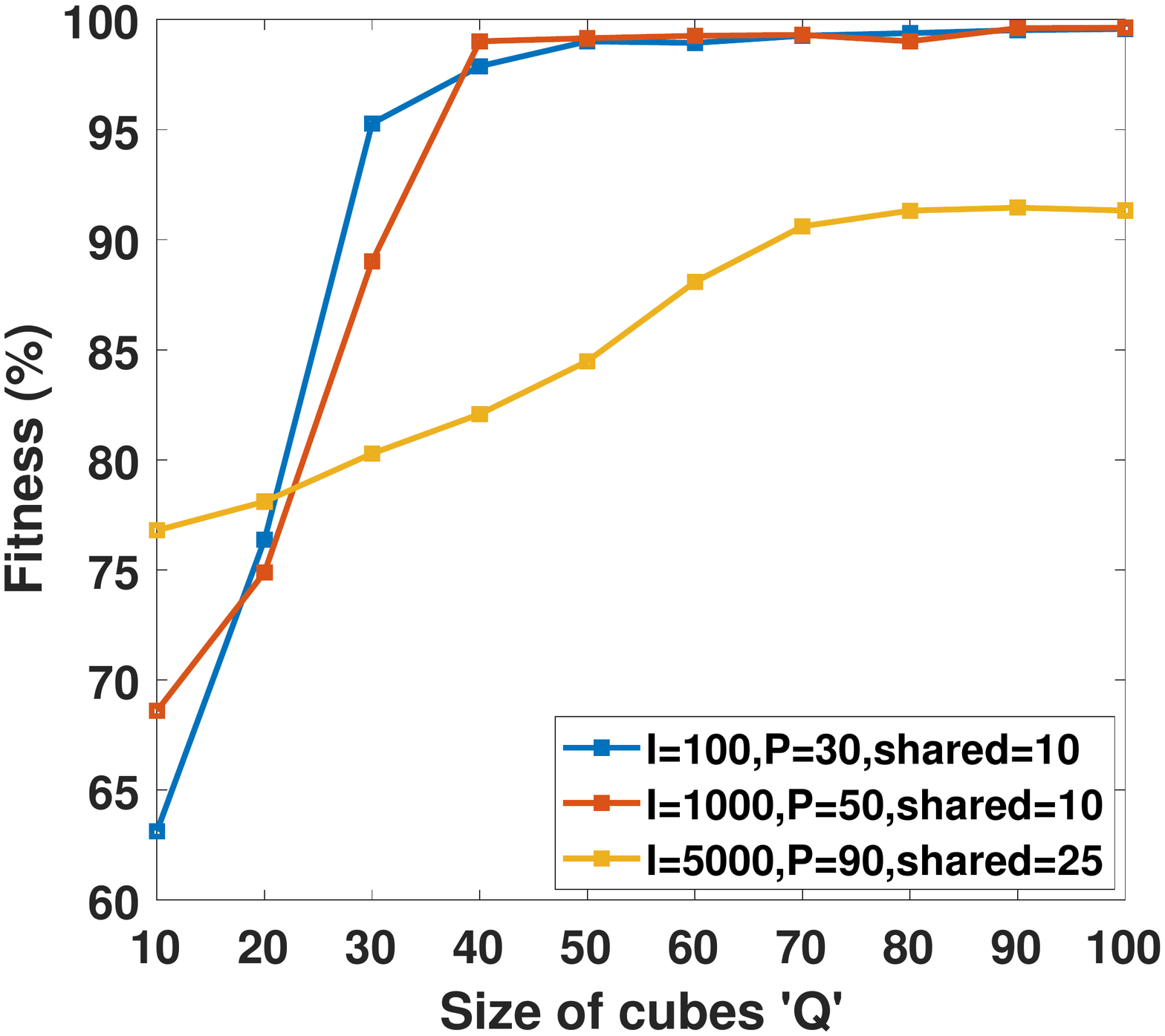}
		\includegraphics[clip,trim=0cm 5cm 0cm 2.5cm,width = 0.25\textwidth]{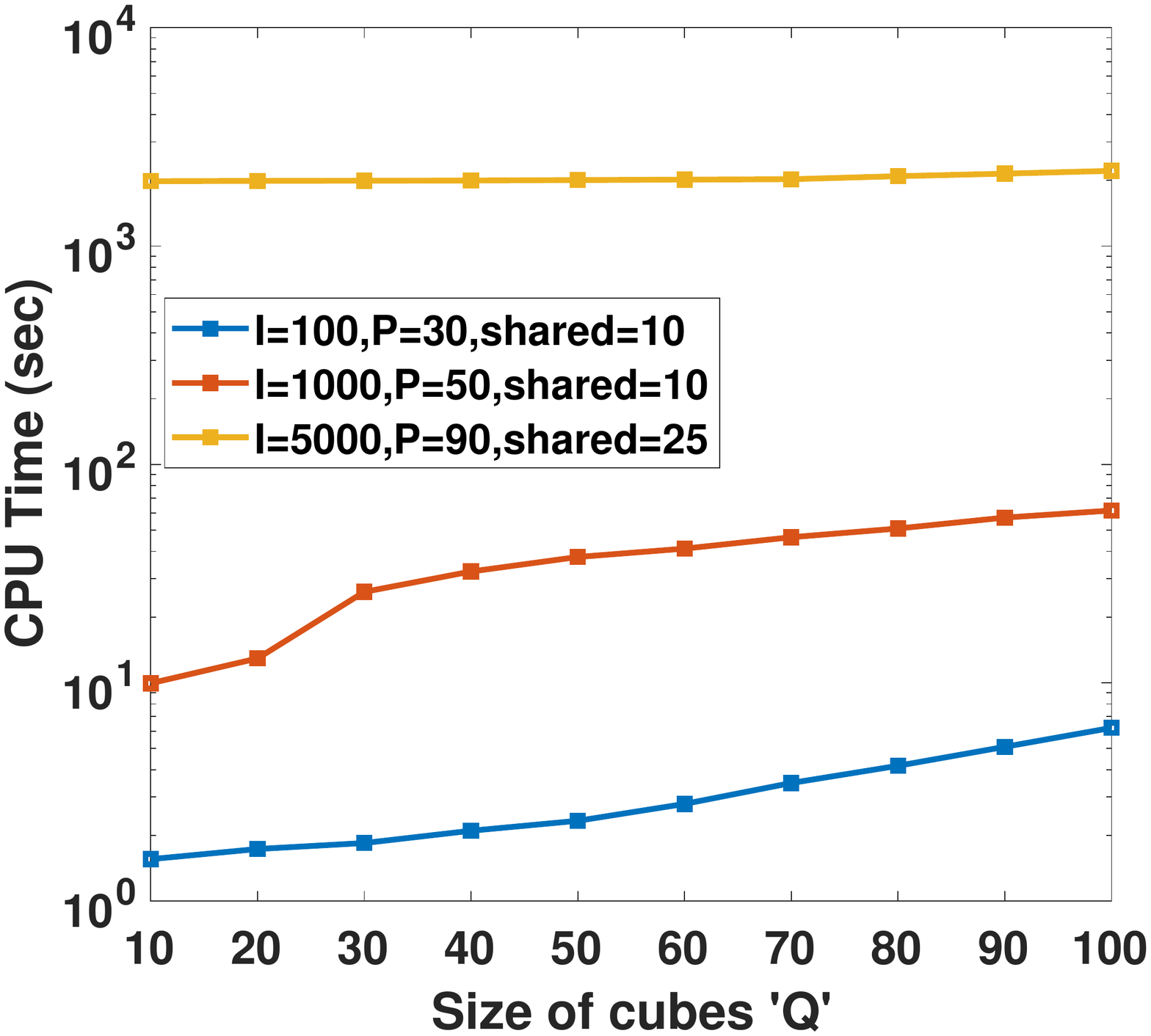}
		\includegraphics[clip,trim=0cm 5cm 0cm 2.5cm,width = 0.25\textwidth]{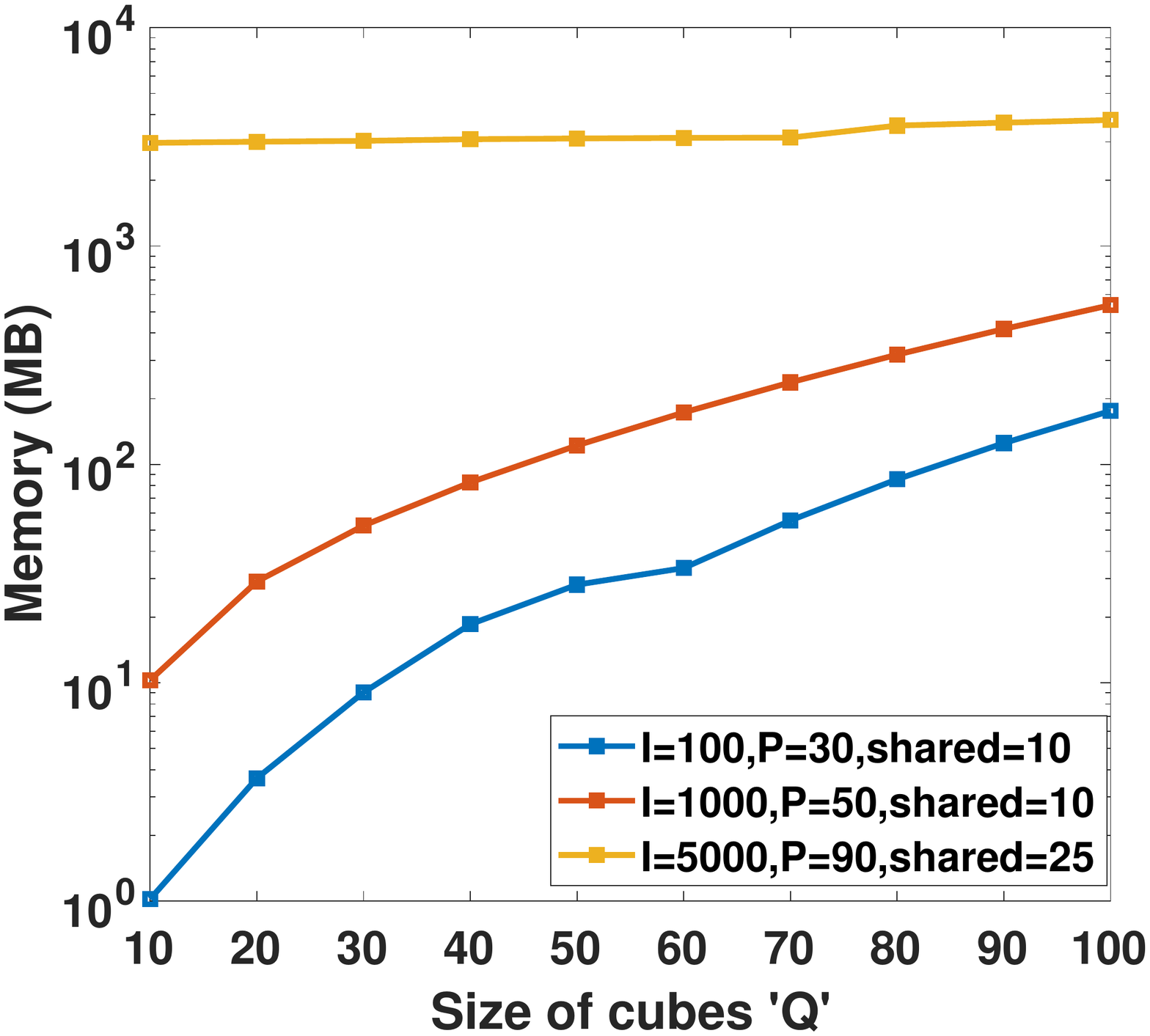}
		\caption{\bf{\method Fitness, CPU Time (sec) and memory used vs. size of compressed tensors 'Q' on different datasets.}}
		\label{fig:sen_Q}
	\end{center}
	 	\vspace{-0.2in}
\end{figure*}

\begin{SCfigure}
	\vspace{-0.4in}
	\includegraphics[clip,trim=0cm 3.8cm 0cm 2.1cm,width = 0.25\textwidth]{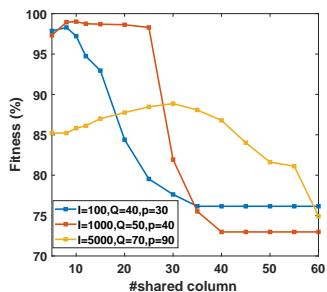}
	\caption{\bf{\method fitness vs. shared columns of compressed tensors '$shared$' on different datasets. It is observed that parameter '$shared$' has negligible effect on CPU time (sec) and memory used(MB).}}
	\label{fig:sen_shared}
\end{SCfigure}
%\vspace{-0.4in}

\hide{
\begin{figure}[!ht]
	\begin{center}
		\includegraphics[clip,trim=1cm 3.8cm 0cm 2.1cm,width = 0.2\textwidth]{FIG/sharedAnalysis.pdf}
		\caption{\bf{\method fitness vs. shared columns of compressed tensors '$shared$' on different datasets. It is observed that parameter '$shared$' has negligible effect on CPU time (sec) and memory used(MB).}}
		\label{fig:sen_shared}
	\end{center}
\end{figure}
}
\textbf{(b) Sensitivity of \textit{Q}}: To evaluate the impact of Q , we fixed other parameters i.e. '$p$' and '$shared$'. We can see that with higher values of the 'Q', Fitness is improved as shown in Figure \ref{fig:sen_Q}. Also It is observed that when equation \ref{equ:para} satisfy, fitness become saturated. Higher the size of compressed cubes, more memory is required to store them. \\

\textbf{(c) Sensitivity of \textit{shared}} :  To evaluate the impact of 'shared' , we fixed other parameters i.e. '$p$' and 'Q'.We observed that this parameter does not have impact on CPU Time (sec) and Memory space(MB). The best fitness is found when $shared \le \frac{Q}{2}$ as shown in figure \ref{fig:sen_shared}. Fitness decreases when $shared \ge \frac{Q}{2}$ because the new compressed cubes completely loses its own structure when joined to old compressed cubes. To retain both old and new structure we choose to keep parameter $shared \le \frac{Q}{2}$ for all experiments.
\hide{
\textbf{(d) Effect of \textit{noise} on decomposition accuracy} :
Additionally, we check the decomposition's quality (fitness) with varying two types of noise level i.e. additive and destructive in data. We increase additive and destructive noise level from 0.1 to 0.5. Figure \ref{fig:noiseAccuracy}(a) shows, the fitness of both methods decreases in proportion to the additive noise level. The fitness of \method is same as compared to base method i.e. Paracomp. Destructive noise makes the tensor factorization harder by making tensor more sparse and introducing noise at the same time. Figure \ref{fig:noiseAccuracy}(b) shows that \method produces as par accurate results as base method.
\begin{figure}[!ht]
	\begin{center}
		\includegraphics[clip,trim=0cm 3.8cm 0cm 2.1cm,width = 0.44\textwidth]{FIG/sharedAnalysis.pdf}
		\includegraphics[clip,trim=0cm 3.8cm 0cm 2.1cm,width = 0.44\textwidth]{FIG/sharedAnalysis.pdf}
		\caption{\bf{\method fitness vs. noise levels. It is observed that with noise the fitness deceases in proportion.}}
		\label{fig:noiseAccuracy}
	\end{center}
	\vspace{-0.5cm}
\end{figure}

\textbf{(e) Effect of \textit{Rank} on decomposition accuracy} : We vary the tensor rank from 5 to 50 with step size of 5. As shown in Figure  \ref{fig:rankAccuracy}, all the evaluation  measures increase in proportion to the rank. This is an expected result since, given a fixed sparsity, the increase in the rank of tensor leads to increased number of non-zeros elements in the input tensor.

\begin{figure}[!ht]
	\begin{center}
		\includegraphics[clip,trim=0cm 3.8cm 0cm 2.1cm,width = 0.44\textwidth]{FIG/sharedAnalysis.pdf}
		\includegraphics[clip,trim=0cm 3.8cm 0cm 2.1cm,width = 0.44\textwidth]{FIG/sharedAnalysis.pdf}
		\caption{\bf{\method fitness, CPU time (sec) and memory used(MB) vs. rank of tensor.}}
		\label{fig:rankAccuracy}
		\end{center}
		\vspace{-0.3in}
	\end{figure}
}

In sum, these observations demonstrate that: 1) a suitable number of cubes and its size i.e. $p,Q$ on compressed tensor could improve the fitness  and result in better tensor decomposition, and 2) For identifiability 'p' must satisfy the condition, $p \geq \max([\frac{(I-shared)}{(Q-shared)} \ \ \frac{J}{Q}  \ \ \frac{K}{Q}])$, to achieve better fitness, lower CPU Time (seconds) and low memory space (MB). This result answers Q4.\\
\subsubsection{\textbf{[Q5] Effectiveness on real world dataset}}
To evaluate effectiveness of our method on real world networks, we use the Foursquare-NYC sequential temporal dataset \cite{yang2014modeling} and American College Football Network (ACFN) \cite{egonetTensors2016} (analysis provided in \textbf{\em{supplementry material}}).\hide{We construct the ACFN tensor data with the player-player interaction to a 115 x 115 grid from using method [] , and considering the time as the third dimension of the tensor. Therefore, each element in the tensor is an integer value that represents the number of interactions between players at a specific moment of time.} Foursquare-NYC dataset includes long-term ($\approx$ 10 months) check-in data in New York city collected from Foursquare from 12 April 2012 to 16 February 2013. % {Our aim is to find the players communities (ground truth communities = 12) changed over time in football dataset and find next top@5 places to visit in NYC per each user. In order to evaluate the effectiveness of our method on football dataset, we compare the ground-truth communities against the communities found by the our method. Figure \ref{fig:realAllcommunities} shows a visualization of the football network over time, with nodes colored according to the observed communities. American football leagues are tightly-knit communities because of very limited matches played across communities. Since these communities generally do not overlap, we perform hard clustering. We find that  "" communities are volatile and players belongs to "" nationality are highly dynamic in forming groups. We observe that \method is able to find relevant communities and also shows the ability to capture the changes in forming those communities in temporal networks. 
\hide{
\begin{figure}[!ht]
	\begin{center}
		\includegraphics[width = 0.23\textwidth]{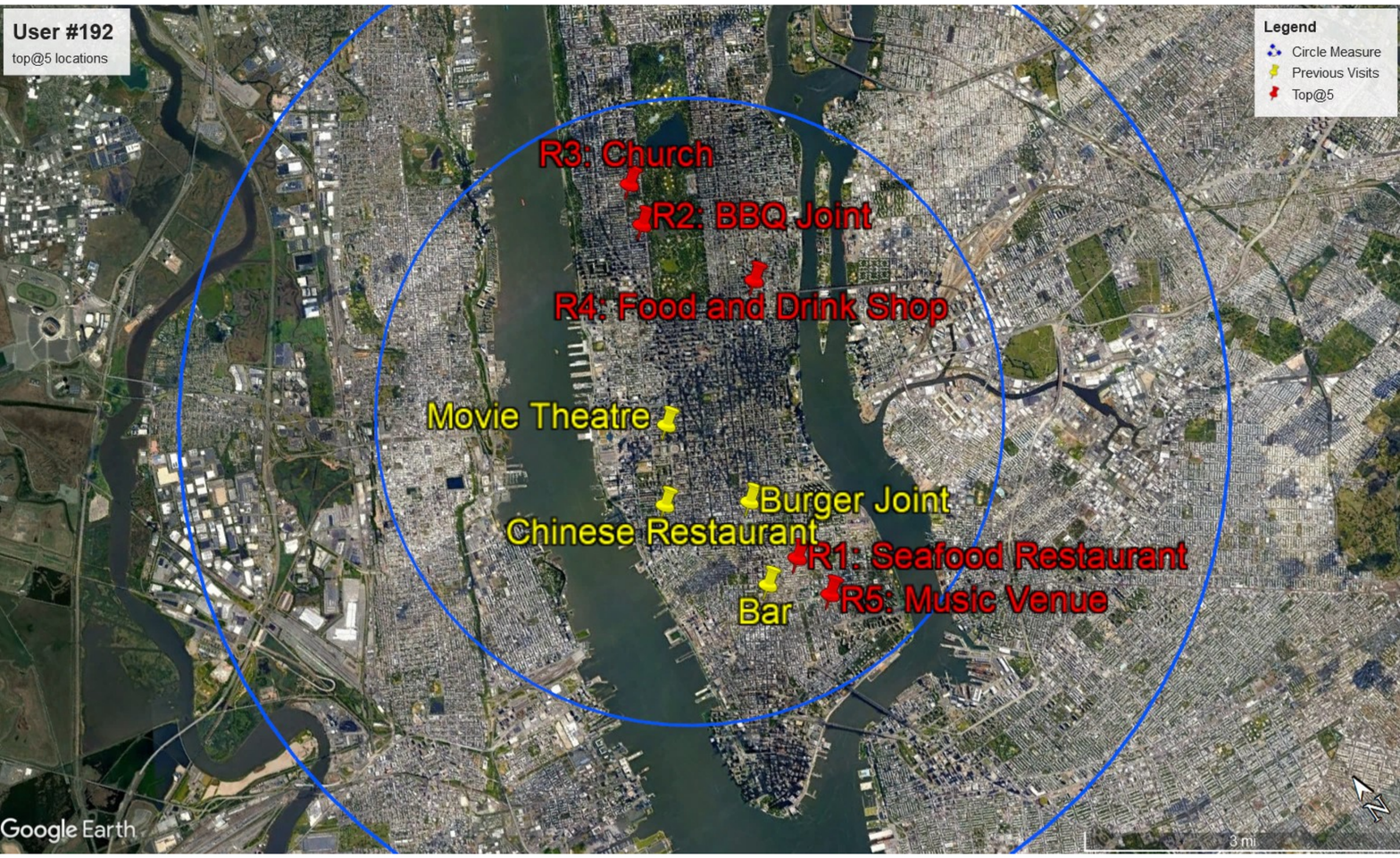}
		\includegraphics[width = 0.23\textwidth]{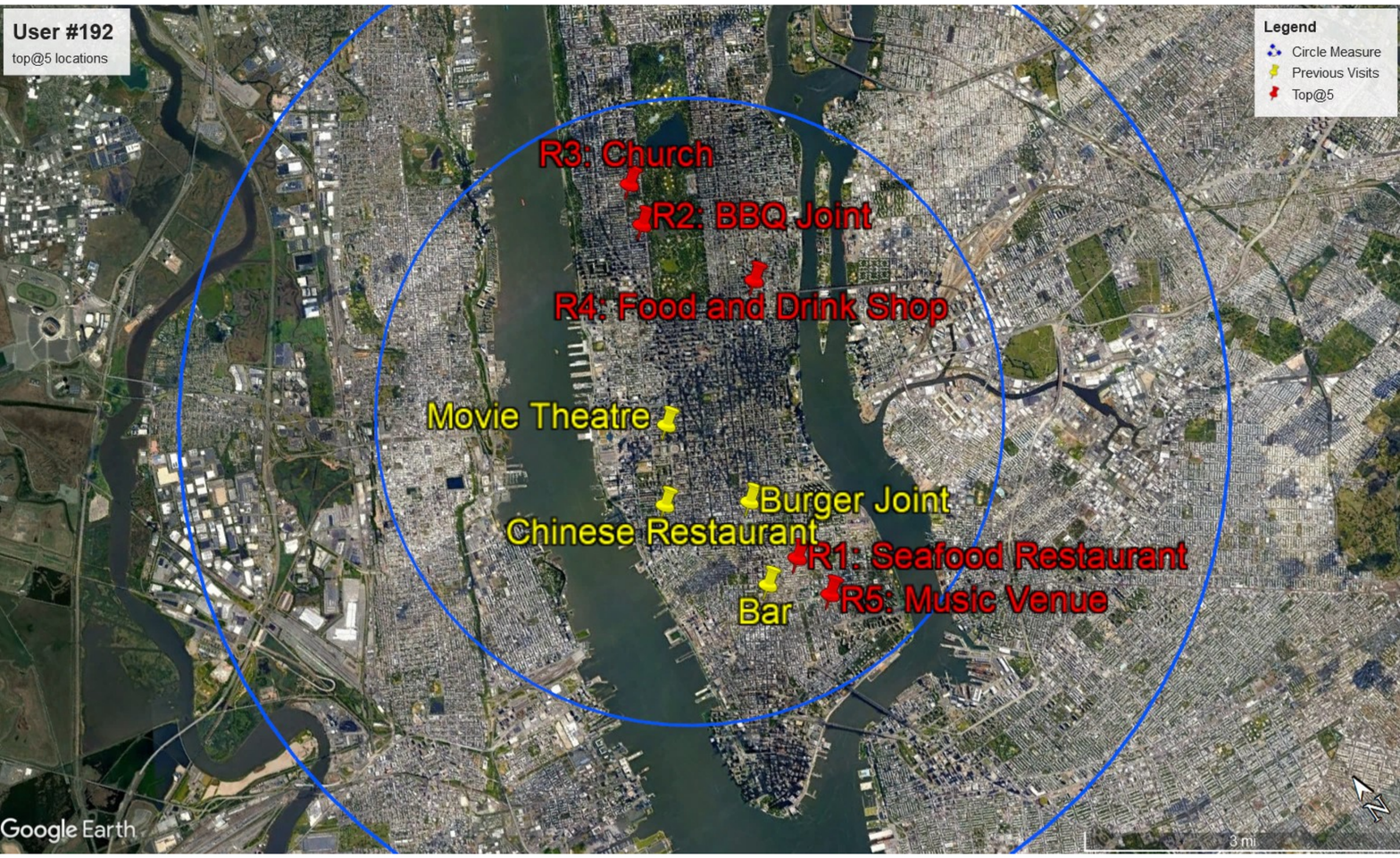}
		\includegraphics[width = 0.23\textwidth]{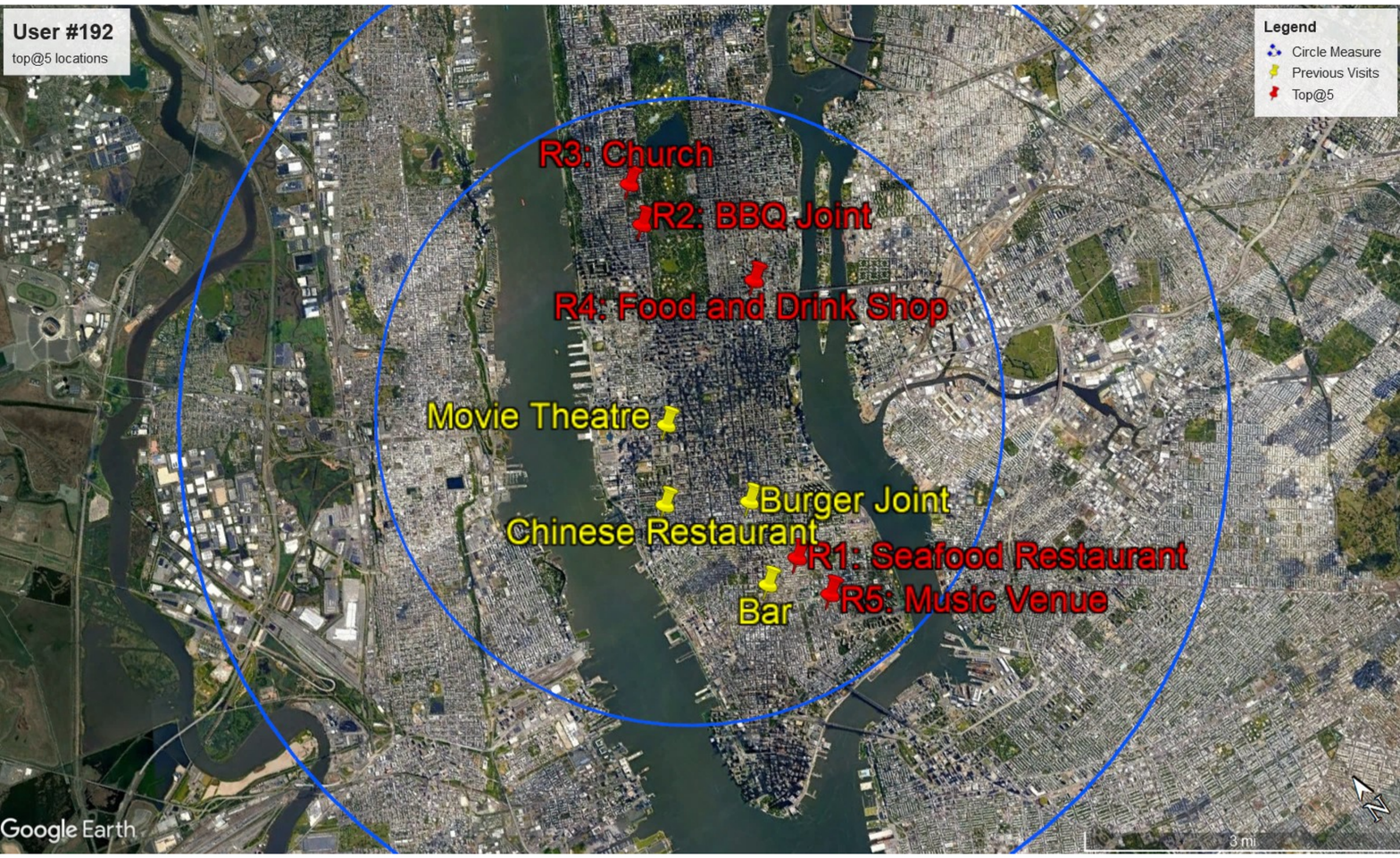}
		\includegraphics[width = 0.23\textwidth]{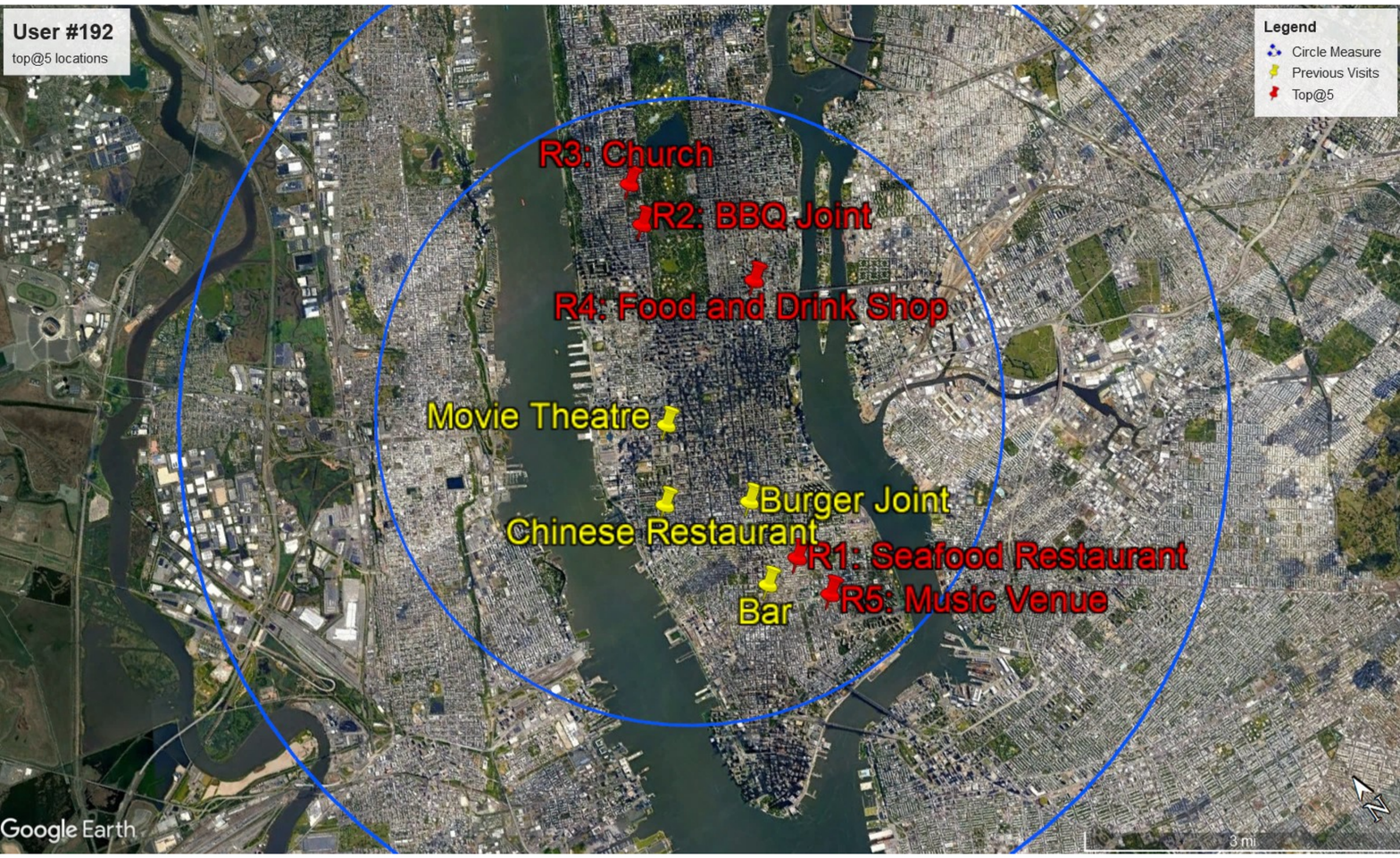}
		\caption{\bf{Visualization of the ground truth communities vs. the identified communities using \method on American College Football Network (ACFN), which has 12 observed players communities.\hide{ \textbf{(a):} Represents the visualization of the network colored with ground truth communities. \textbf{(b):} Shows the visualization of the network colored with predicted communities at time $ \frac{1}{3} T$, where $T$ is total time stamps. \textbf{(c):} Shows the visualization of the network colored with predicted communities at time $ \frac{2}{3} T$. \textbf{(d):} Shows the visualization of the network colored with predicted communities at time $T$.} We see that reconstructed views using \method helps to identify the communities changing over time.}}
		\label{fig:realAllcommunities}
	\end{center}
\vspace{-0.3in}
\end{figure}
}
The tensor data is structured as [user (1k), Point of Interest (40k), time (310 days)] and each element in the tensor represents the total time spent by user for that visit. Our aim is to find next top@5 places to visit in NYC per user. We decompose the tensor data into batches of 7 days and using rank = $15$ estimated by AutoTen \cite{papalexakis2016automatic}. For evaluation, we reconstruct the complete tensor from factor matrices and mask the known POIs in the tensor and then rank the locations for each user. Since we do not have human supplied relevance rankings for this dataset, we choose to visualize the most significant factor (locations) using maps provided by Google. If the top ranked places are with-in 5 miles radius of user's previous places visited, then we consider the decomposition is effective. In the Figure \ref{fig:realAllplaces}(a), the five red markers corresponds to the five highest values of the factor. These locations correspond to well-known area in NYC : Brooklyn Bridge , Square garden , Theater District and Museum of Art. The high density of activities (green points) verifies their popularity. Figure \ref{fig:realAllplaces}(b,c) shows the top@5 results for users {\#}192 and user {\#}902, the red marker shows the next locations to visit and yellow marker shows the previous visited locations. More interestingly, we can see that user {\#}192 visited coffee shops and restaurants most of the time food, top@5 ranked locations are also either restaurants or food \& drink shops. Similarity, user {\#}902, most visited places are  Fitness center, top@5 ranked locations are park, playground and Stadium. Both case studies shows the effectiveness of the decomposition and confirms that the \method can be used for various types of data analysis and this answers \textbf{Q5}.
\hide{
\begin{figure}[!ht]
	
	\begin{center}
		\includegraphics[width = 0.43\textwidth]{casestudy/famus500top5_2.PNG}
		\caption{\bf{\method's five highest values of the factor are represented as red markers. \hide{These locations are popular in New York and shows that decomposition is effective and accurate. We use Google Earth software to plot the POIs. The markers are pinned to corresponding latitude and longitude of POI given in NYC-dataset.} }}
		\label{fig:realAllplaces}
	\end{center}
	\vspace{-0.3in}
\end{figure}
\begin{figure}[!ht]
	
	\begin{center}
		\includegraphics[width = 0.45\textwidth]{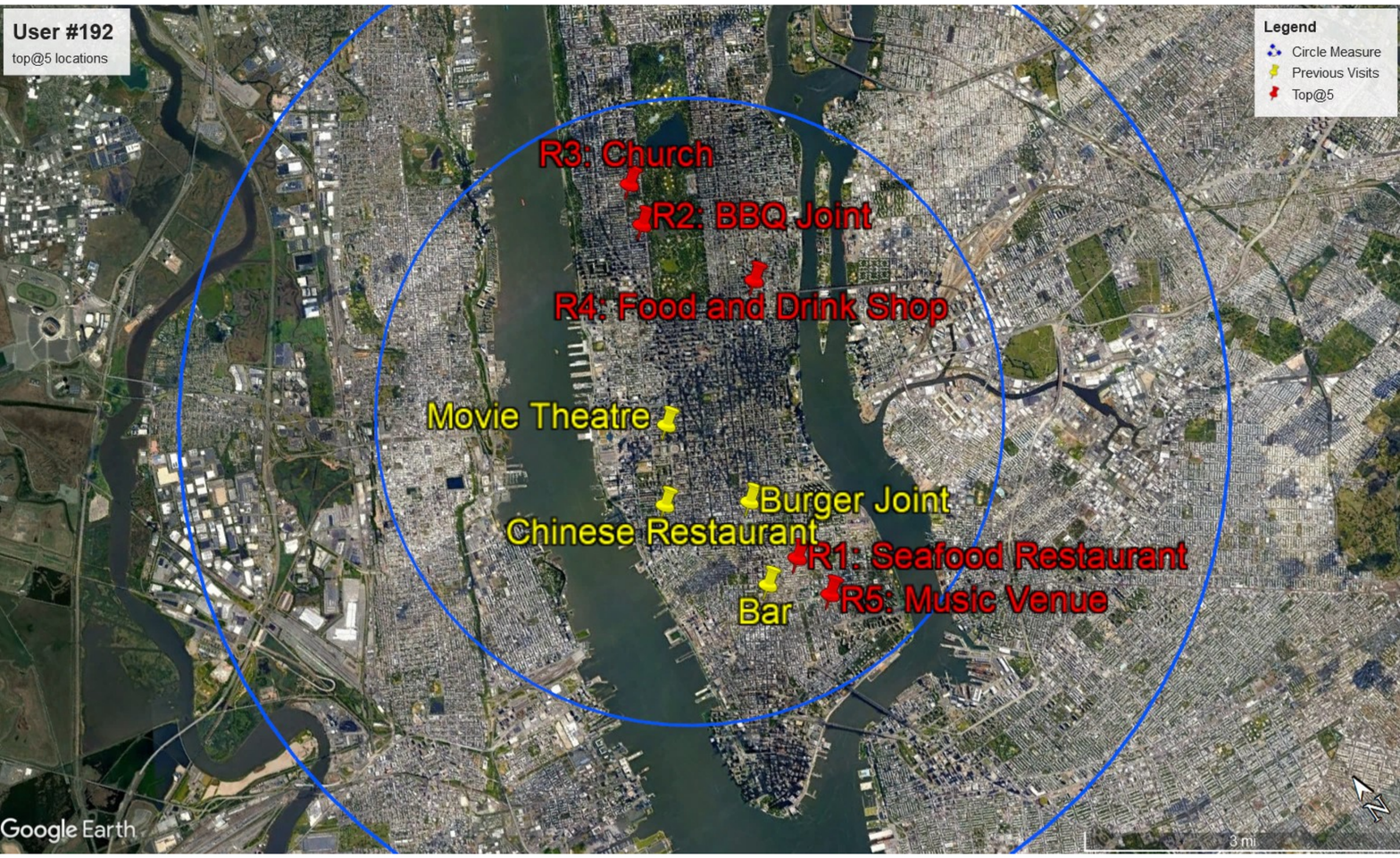}
		\includegraphics[width = 0.45\textwidth]{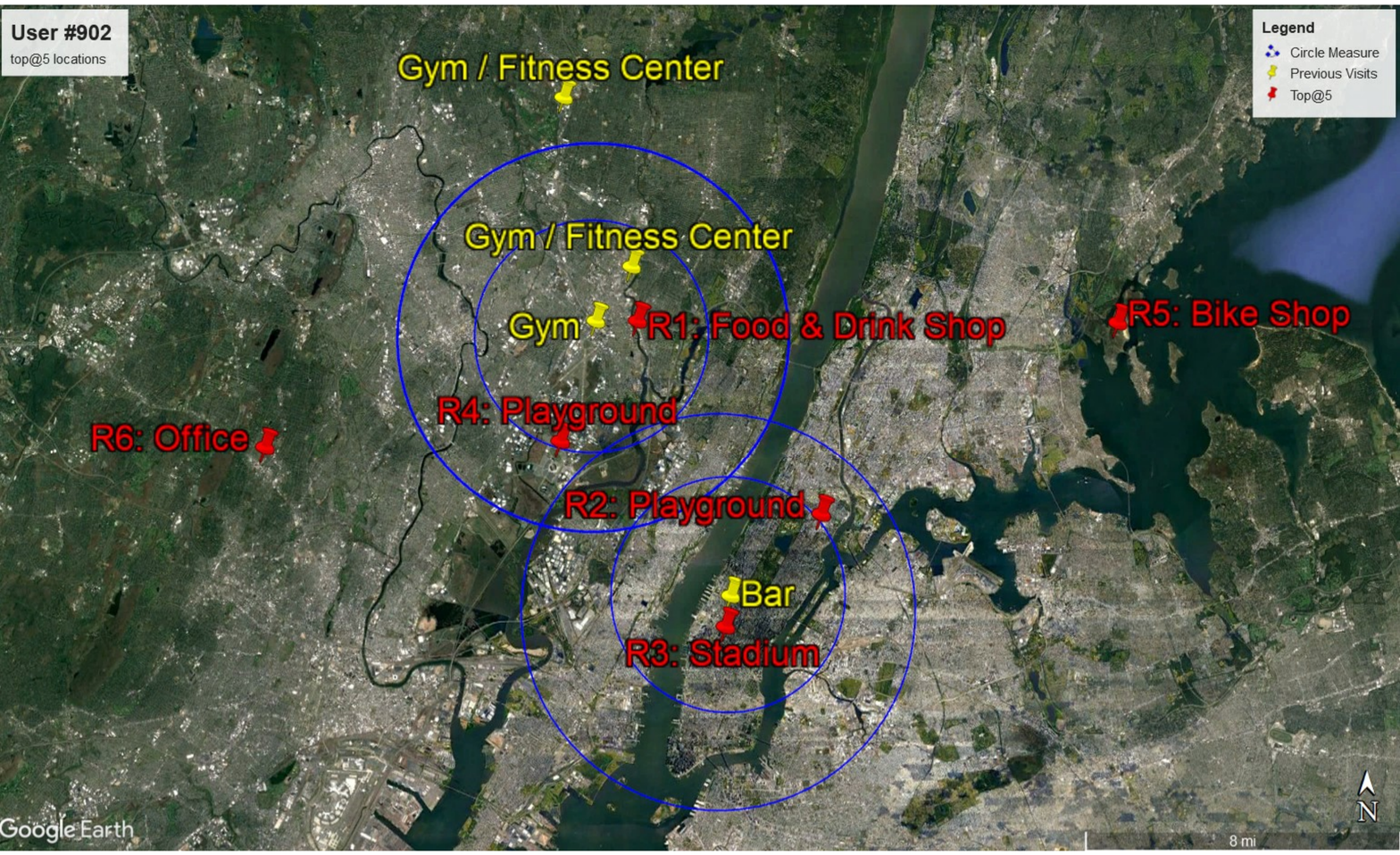}
		\caption{\bf{Visualization of the top@5 POIs of the \textbf{user{\#}192 and user{\#}902} obtained from reconstructed tensor using factor matrices. The yellow markers are user's previous visited POIs and red markers are recommended POIs.\hide{ \textbf{(a):} Visualizes the top@5 POIs for user {\#} 192 \textbf{(b):}Visualizes the top@5 POIs for user {\#} 902. \method is able to find the new POIs with-in 3 (inner circle)-5 (outer circle) mile radius of previous visited places  or POIs. We observe same results for about 70\% users and rest 30\% users have either have extremely less check-in data points or \method recommended POIs beyond 10 miles.}  }}
		\label{fig:realuser902}
	\end{center}
\vspace{-0.1in}
\end{figure}
}
\begin{figure*}[!ht]
		\begin{center}
			\includegraphics[clip,trim=0cm 1cm 0cm 1cm,width = 0.32\textwidth]{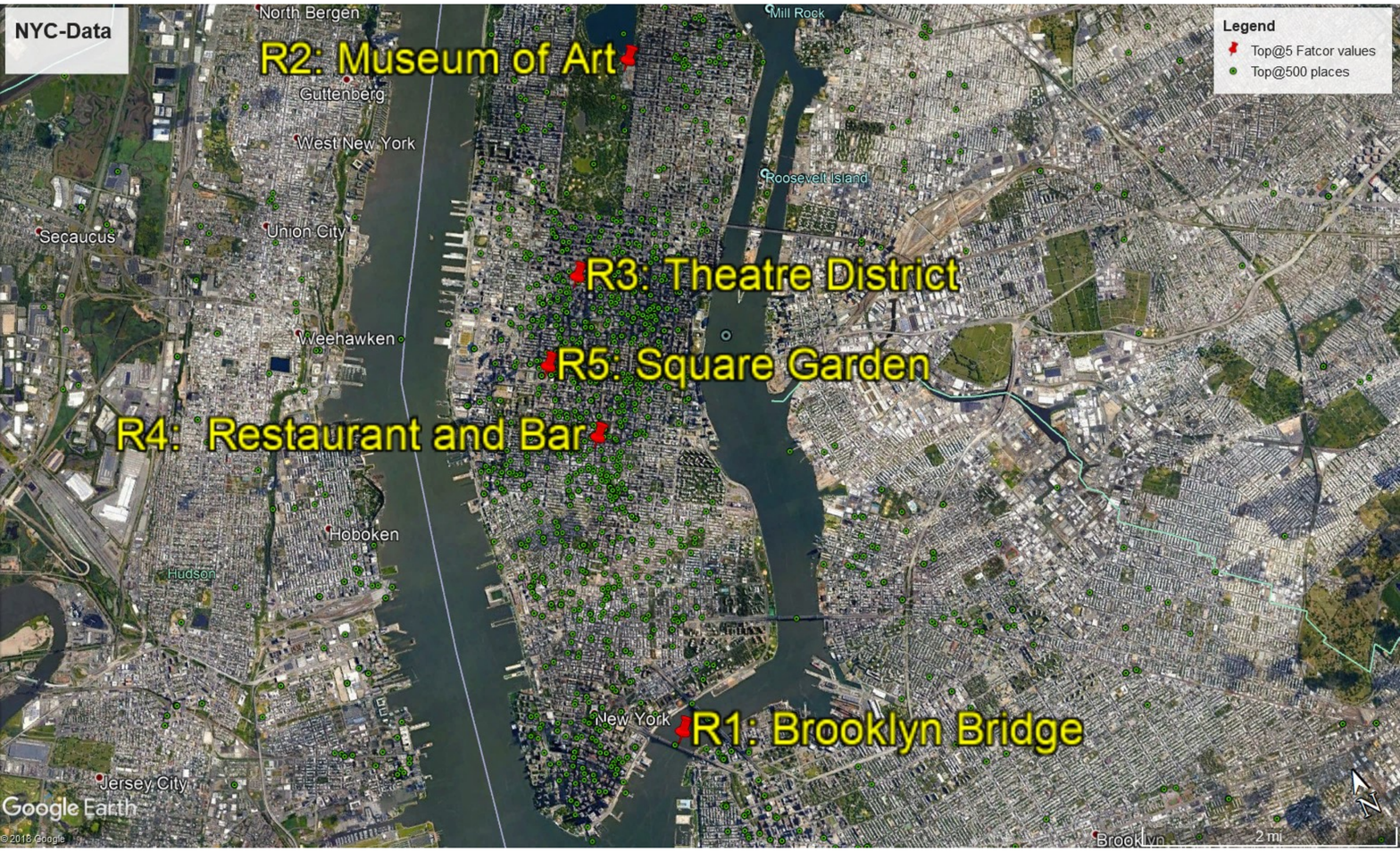}
			\includegraphics[clip,trim=0cm 1cm 0cm 1cm,width = 0.32\textwidth]{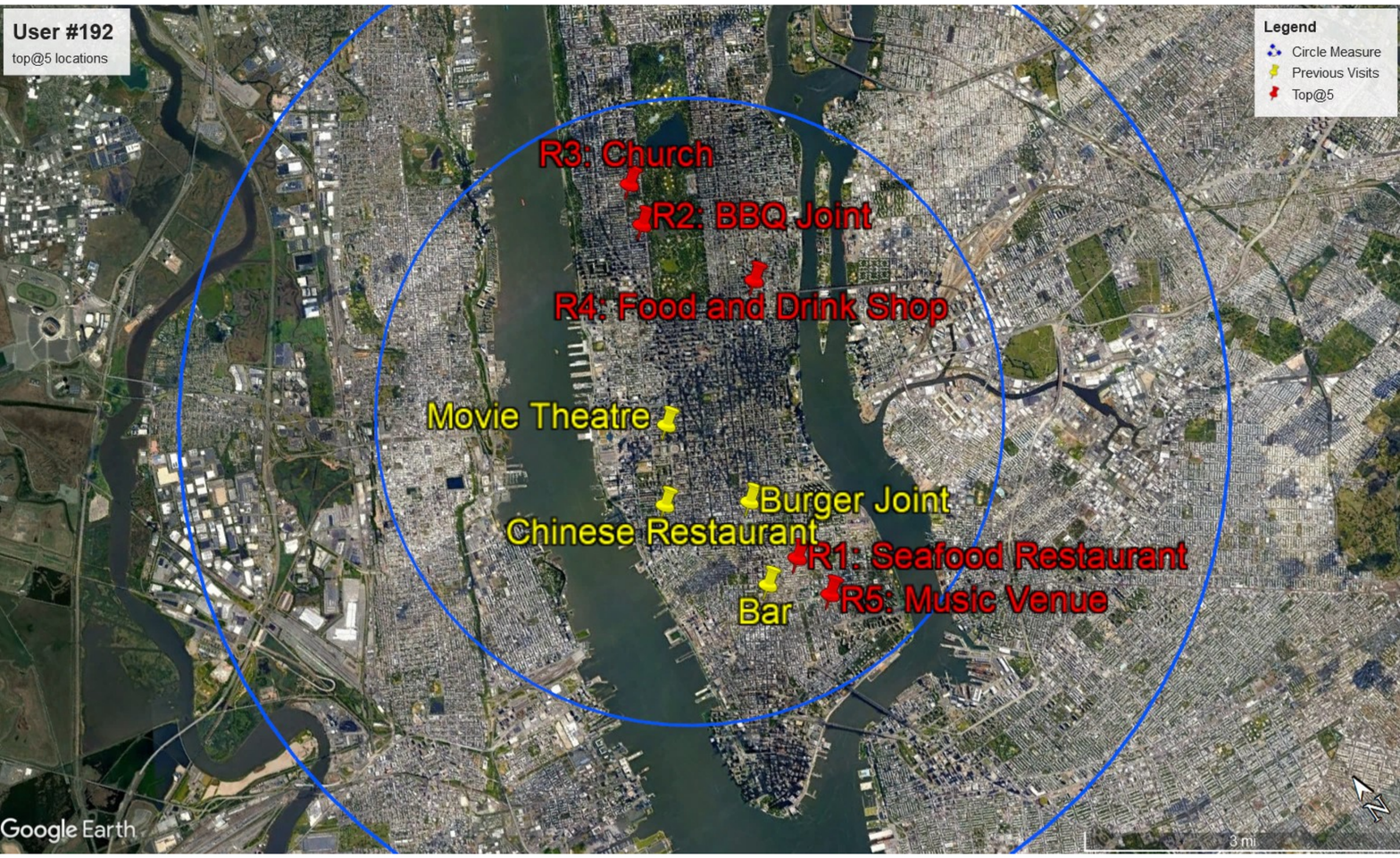}
			\includegraphics[clip,trim=0cm 1cm 0cm 1cm,width = 0.32\textwidth]{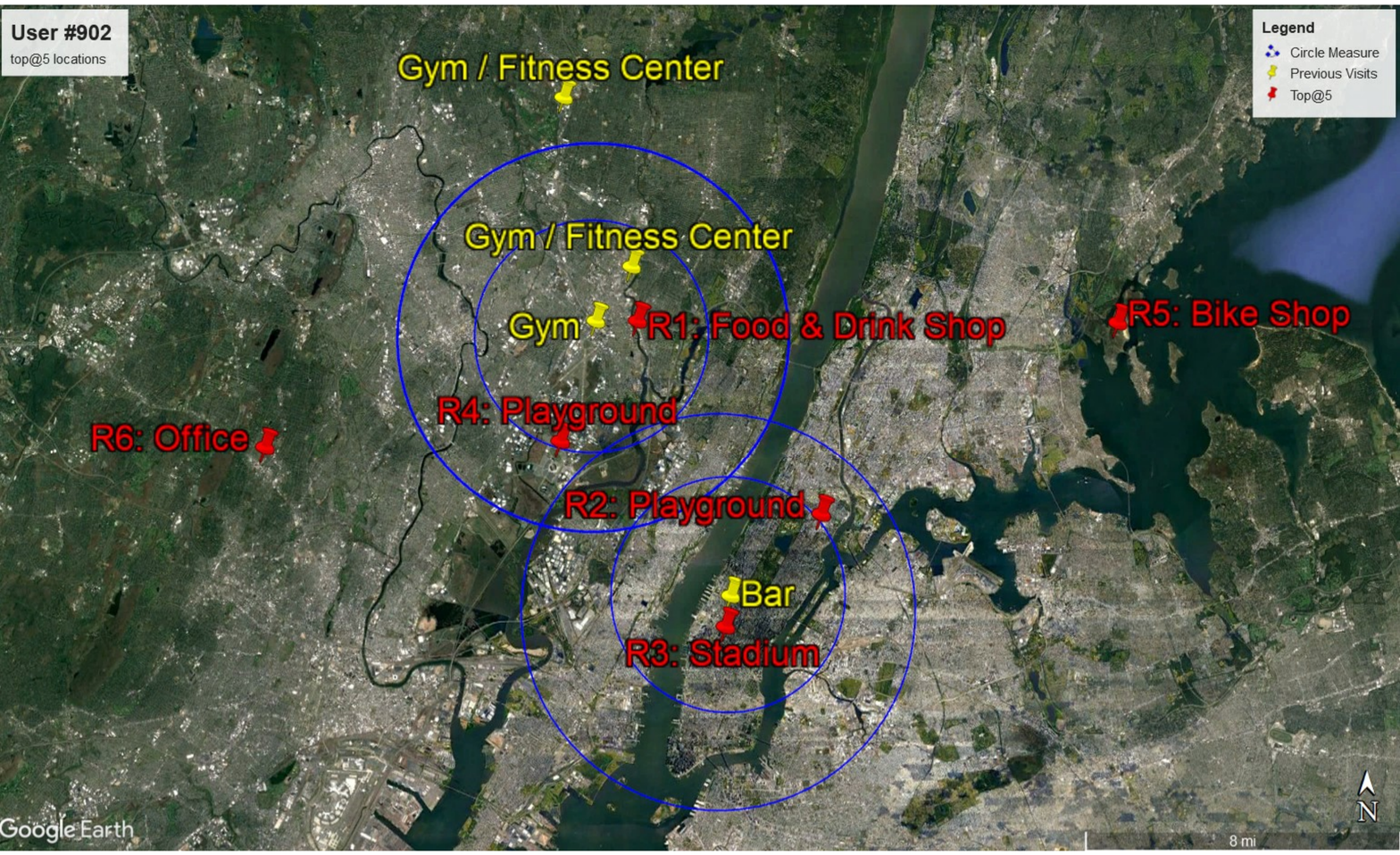}
			\caption{\bf{(a)\method's five highest values of the factor are represented as red markers. (b,c) Visualization of the top@5 POIs of the \textbf{user{\#}192 and user{\#}902} obtained from reconstructed tensor using factor matrices. The yellow markers are user's previous visited POIs and red markers are recommended POIs. }}
			\label{fig:realAllplaces}
		\end{center}
		\vspace{-0.3in}
	\end{figure*}

%% file: 050related.tex
\section{Related Work}
\label{sec:related}
In this section, we provide review of the work related to our algorithm. At large,  incremental tensor methods in the literature can be categorized into two main categories as described below:

\noindent{\bf Tensor Decomposition}: Phan \textit{el at.} \cite{phan2011parafac} had purposed a theoretic method namely GridTF to large-scale tensors decomposition based on CP's basic mathematical theory to get sub-tensors and join the output of all decompositions to achieve final factor matrices. Sidiropoulos \textit{el at.}\cite{nion2009adaptive}, proposed algorithm that focus on CP decomposition namely RLST (Recursive Least Squares Tracking), which recursively update the factors by minimizing the mean squared error. In 2014, Sidiropoulos \textit{el at.} \cite{sidiropoulos2014parallel} , proposed a parallel algorithm for low-rank tensor decomposition that is suitable for large tensors. The Zhou, \textit{el at.} \cite{zhou2016accelerating} describes an online CP decomposition method, where the latent components are updated for incoming data. The most related work to ours was proposed by \cite{gujral2017sambaten} which is sampling-based batch incremental tensor decomposition algorithm. These state-of-the-art techniques focus on only fast computation but not effective memory usage. Besides CP decomposition, tucker decomposition methods\cite{SunITA,papadimitriou2005streaming} were also introduced. Some of these methods were not only able to handle data increasing in one-mode, but also have solution for multiple-mode updates using methods such as incremental SVD \cite{fanaee2015multi}. Latest line of work is introduced in \cite{austin2016parallel} i.e TuckerMPI to find inherent low-dimensional multi-linear structure, achieving high compression ratios. Tucker is mostly focused on recovering subspaces of the tensor, rather than latent factors, whereas our focus is on the CP decomposition which is more suitable for exploratory analysis.

\noindent{\bf Tensor Completion}:
Another field of study is tensor completion, where real-world large-scale datasets are considered incomplete. In literature, wide range of methods have been proposed based on online tensor imputation\cite{mardani2015subspace} and  tensor completion with auxiliary information\cite{narita2011tensor,acar2011all}. The most recent method in this line of work is by Qingquan \textit{el at.}\cite{song2017multi},  who proposed a low-rank tensor completion with general multi-aspect streaming patterns, based on block partitioning of the tensor. However, these approaches cannot be directly applied when new batches of data arrived. This provides us a good starting reference for further research.

%% file: 060conclusions.tex
\section{Conclusions}
\label{sec:conclusions}
In this work, we focus on online tensor decomposition problem and proposed a novel compression based \method framework. The proposed framework effectively identify the low rank latent factors of compressed replicas of incoming slice(s) to achieve online tensor decompositions. To further enhance the capability, we also tailor our general framework towards  higher-order online tensors. Through experiments, we empirically validate its effectiveness and accuracy and we demonstrate its memory efficiency and scalability by outperforming state-of-the-art approaches (40-200 \% better). Regardless, future work will focus on investigating different tensor decomposition methods and incorporating various tensor mining methods into our framework.

%% file: 070Supplement.tex
\section{Supplementary Materials}
\label{sec: supplementary}
\subsection{\textbf{Extending to Higher-Order Tensors} }
We now show how our approach is extended to higher-order cases. Consider N-mode tensor $\tensor{X}_{old} \in \mathbb{R}^{I_1 \times I_2 \times \dots \times I_{N-1} \times t_{old}}$. The factor matrices are $(\mathbf{A}^{(1)}_{old}, \mathbf{A}^{(2)}_{old},\dots, \mathbf{A}^{(N-1)}_{old}, \mathbf{A}^{(T_1)}_{old})$ for CP decomposition with $N^{th}$ mode as new incoming data. A new tensor $\tensor{X}_{new} \in \mathbb{R}^{I_1 \times I_2 \times \dots \times I_{N-1} \times t_{new}}$ is added to $\tensor{X}_{old}$ to form new tensor of $\mathbb{R}^{I_1 \times I_2 \times \dots  \times I_{N-1} \times T}$ where $T=t_{old}+t_{new}$. In addition, sets of {\em compression matrices} for old data are \{$\mathbf{U}_p^{(1)},\mathbf{U}_p^{(2)},\dots, \mathbf{U}_p^{(N-1)}, \mathbf{U}_p^{(T)}$\} and for new data it is \{$\mathbf{U}_p^{'(1)},\mathbf{U}_p^{'(2)},\dots, \mathbf{U}_p^{'(N-1)}, \mathbf{U}_p^{'(T)}$\} for {\em p} number of summaries.

Each compression matrices are converted into compressed cubes i.e. for $\tensor{X}_{old}$ compressed cube is of dimension $\tensor{Y}_{p} \in \mathbb{R}^{Q^{(1)} \times Q^{(2)}  \dots \times Q^{(N-1)}  \times Q^{(N)} }$ and same follows for $\tensor{X}_{new}$. The updated summaries are computed using $\tensor{X}_{p}=\tensor{Y}_{p}+\tensor{Z}_{p}$ s.t. $\tensor{X}_{p}  \in \mathbb{R}^{Q^{(1)} \times Q^{(2)}  \dots \times Q^{(N-1)}  \times Q^{(N)} }$. After CP decomposition of  $\tensor{X}_{p}$, factor matrices and random compressed matrices are stacked as : 
\begin{multline*}
(\widetilde{\mathbf{A}}_{s}^{(1)},\dots, \widetilde{\mathbf{A}}_{s}^{(N-1)} , \widetilde{\mathbf{A}}_{s}^{(N)}) \leftarrow \Pi\big[(\widetilde{A}_{s(i)}^{(1)}, \dots, \widetilde{\mathbf{A}}_{s(i)}^{(N-1)},\widetilde{\mathbf{A}}_{s(i)}^{(N)} ) {;} \\ 
\ \ (\widetilde{\mathbf{A}}_{s(i+1)}^{(1)}, \widetilde{\mathbf{A}}_{s(i+1)}^{(2)} ,\dots, \widetilde{\mathbf{A}}_{s(i+1)}^{(N-1)},\widetilde{\mathbf{A}}_{s(i+1)}^{(N)} )\big] , i \in (1,p-1)
\end{multline*}
\begin{equation}
\ssmall
\& \  \ (\widetilde{\mathbf{P}}^{(1)},\widetilde{\mathbf{P}}^{(2)},\dots, \widetilde{\mathbf{P}}^{(N-1)} , \widetilde{\mathbf{\mathbf{P}}}^{(N)}) = \begin{bmatrix}
\widetilde{\mathbf{P}}_{(1)}^{(1)},\widetilde{\mathbf{P}}_{(1)}^{(2)}, \dots, \widetilde{\mathbf{P}}_{(1)}^{(N-1)} , \widetilde{\mathbf{P}}_{(1)}^{(N)}\\
\widetilde{\mathbf{P}}_{(2)}^{(1)},\widetilde{\mathbf{P}}_{(2)}^{(2)}, \dots, \widetilde{\mathbf{P}}_{(2)}^{(N-1)} , \widetilde{\mathbf{P}}_{(2)}^{(N)}\\
\vdots\\
\widetilde{\mathbf{P}}_{(p)}^{(1)},\widetilde{\mathbf{P}}_{(p)}^{(2)}, \dots, \widetilde{\mathbf{P}}_{(p)}^{(N-1)} , \widetilde{\mathbf{P}}_{(p)}^{(N)}
\end{bmatrix} 
\end{equation}
Finally, the update rule of each non-temporal mode $\in (1,N-1)$ and temporal mode $\in (N)$ is :
\begin{multline*}
(\mathbf{A}^{(1)} ,\dots ,\mathbf{A}^{(N-1)}, \mathbf{A}^{N}) \leftarrow
(\widetilde{\mathbf{P}}^{(1)^{-1}}*\widetilde{\mathbf{A}}_{s}^{(1)}, \dots ,
\end{multline*}
\begin{equation} 
\widetilde{\mathbf{P}}^{(N-1)^{-1}}*\widetilde{\mathbf{A}}_{s}^{(N-1)}, \big[ \mathbf{A}_{old}^{(N)}; \widetilde{\mathbf{P}}^{(N)^{-1}}*\widetilde{\mathbf{A}}_{s}^{(N)}) \big]
\end{equation}
\hide{
	$$
	\begin{bmatrix}
	A^{(1)}\\
	A^{(2)}\\
	\vdots\\
	A^{(N-1)}\\
	A^{N}
	\end{bmatrix} 
	\leftarrow
	\begin{bmatrix}
	\widetilde{P}^{(1)-1}*\widetilde{A}_{s}^{(1)}\\
	\widetilde{P}^{(2)-1}*\widetilde{A}_{s}^{(2)}\\
	\vdots\\
	\widetilde{P}^{(N-1)-1}*\widetilde{A}_{s}^{(N-1)}\\
	\widetilde{P}^{(N)-1}*\widetilde{A}_{s}^{(N)}
	\end{bmatrix} 
	$$
}
\subsection{\textbf{Necessary characteristics for uniqueness}}
As we mention above, \method is able to identify the solution of the online CP decomposition, as long as the parallel CP decompositions on the compressed tensors are also identifiable. Empirically, we observe that if the decomposition\hide{was} of a given data that has \hide{have} exact or near-trilinear structure (or multilinear in the general case), i.e. obeying the low-rank CP model with some additive noise, \method is able to successfully, accurately, and using much fewer resources than state-of-the-art, track the online decomposition. On the other hand, when given data that do not have a low trilinear rank, the results were of lower quality. This observation is of potential interest in exploratory analysis, where we do not know 1) the (low) rank of the data, and 2) whether the data have low rank to begin with (we note that this is an extremely hard problem, out of the scope of this paper, but we refer the interested reader to previous studies \cite{wang2018low,papalexakis2016automatic} for an overview). If \method provides a good approximation, this indirectly signifies that the data have appropriate trilinear structure, thus CP is a fitting tool for analysis. If, however, the results are poor, this may indicate that we need to reconsider the particular rank used, or even analyzing the data using CP in the first place. We reserve further investigations of what this observation implies for future work.
%%%%%%%%%%%%%%%%%%%%%%%%%%%%%%%%%%%%%%%%%%%%%%%%%%%%%%%%%%%%%%%
%%%%%%%%%%%%%% Algorithm %%%%%%%%%%%%%%%%%%%%%%%%%%%%%%%%%%%%%%
%%%%%%%%%%%%%%%%%%%%%%%%%%%%%%%%%%%%%%%%%%%%%%%%%%%%%%%%%%%%%%%

\begin{center}
	%\vspace{-0.25in}
	\begin{algorithm} [!htp]
		\small
		\caption{\method for incremental 3-mode tensor decomposition}
		\begin{algorithmic}
			\REQUIRE $\tensor{X}_{new} \in \mathbb{R}^{I \times J \times K_{(n+1)\dots m}}$,summary $ \tensor{Y}_i \in \mathbb{R}^{Q \times Q \times Q}$,  R,p,Q, shared S.
			
			\ENSURE Factor matrices $\mathbf{A}, \mathbf{B}, \mathbf{C}$ of size $(I \times R)$, $(J \times R)$ and $(K_{1\dots n,(n+1) \dots m} \times R)$.
			
			%\STATE $Y_i \leftarrow \{(U_o(:,[shared,:])_{i},V_o(:,[shared,:])_i, W_o(:,[shared,:])_i)\}$ ,\ \ $ Y_i \in \mathbb{R}^{Q \times Q \times Q}$,\ \ $i \in (1,p)$, for $\tensor{X}_{old}$ \\
			\While{new slice(s) coming} 
			{
				\STATE $\tensor{Z}_i \leftarrow \{(\mathbf{U}_i^{'},\mathbf{V}_i^{'}, \mathbf{W}_i^{'})\}$ ,\ \ $ \tensor{Z}_i \in \mathbb{R}^{Q \times Q \times Q}$,\ \ $i \in (1,p)$  \\
				\STATE $\tensor{X}_i\leftarrow \tensor{Y}_i\oplus \tensor{Z}_i$ , \ \ $ \tensor{X}_i \in \mathbb{R}^{Q \times Q \times Q}$,\ \ $i \in (1,p)$ 
				\STATE $(\widetilde{\mathbf{A}}_{s(i)},\widetilde{\mathbf{B}}_{s(i)},\widetilde{\mathbf{C}}_{s(i)}) \leftarrow CP(\tensor{X}_i,R),\ \ i \in (1,p)$ 
				\STATE $(\widetilde{\mathbf{P}}_{a(i)},\widetilde{\mathbf{P}}_{b(i)},\widetilde{\mathbf{P}}_{c(i)})\leftarrow \{(\mathbf{U}^{'}(i,[S,:],:)^T,\mathbf{V}^{'}(i,[S,:],:)^T,\mathbf{W}^{'}(i,[S,:],:)^T)\},\ \ i \in (1,p)$ 
				\FOR{$i \leftarrow 1$ to $p-1$}
				\STATE $(\widetilde{\mathbf{A}}_{s},\widetilde{\mathbf{B}}_{s},\widetilde{\mathbf{C}}_{s})\leftarrow \Pi\big[(\widetilde{\mathbf{A}}_{s(i)}, \widetilde{\mathbf{B}}_{s(i)} , \widetilde{\mathbf{C}}_{s(i)}) \ \  {;} \ \ (\mathbf{A}_{s(i+1)}, \mathbf{B}_{s(i+1)} , \mathbf{C}_{s(i+1)})\big]$
				\STATE $(\widetilde{\mathbf{P}}_{a},\widetilde{\mathbf{P}}_{b},\widetilde{\mathbf{P}}_{c}) \leftarrow \big[(\widetilde{\mathbf{P}}_{a(i)},\widetilde{\mathbf{P}}_{b(i)},\widetilde{\mathbf{P}}_{c(i)}) \ \  {;} \ \ (\widetilde{\mathbf{P}}_{a(i+1)},\widetilde{\mathbf{P}}_{b(i+1)},\widetilde{\mathbf{P}}_{c(i+1)})\big]$ 
				
				\ENDFOR
				\STATE $\mathbf{A} \leftarrow \widetilde{\mathbf{P}}_a^{-1}*\widetilde{\mathbf{A}}_s$ ; $\mathbf{B} \leftarrow \widetilde{\mathbf{P}}_b^{-1}*\widetilde{\mathbf{B}}_s$ ;   $\mathbf{C} \leftarrow [\mathbf{C}_{old}; \widetilde{\mathbf{P}}_c^{-1}*\widetilde{\mathbf{C}}_s]$
			}
			\RETURN ($\mathbf{A},\mathbf{B},\mathbf{C}$)
		\end{algorithmic}
		\label{alg:method}
	\end{algorithm}
	%	\vspace{-0.1in}
\end{center}

\begin{figure*}[!ht]
	\begin{center}
		\includegraphics[clip,trim=1cm 0.1cm 0.5cm 1cm,width = 1\textwidth]{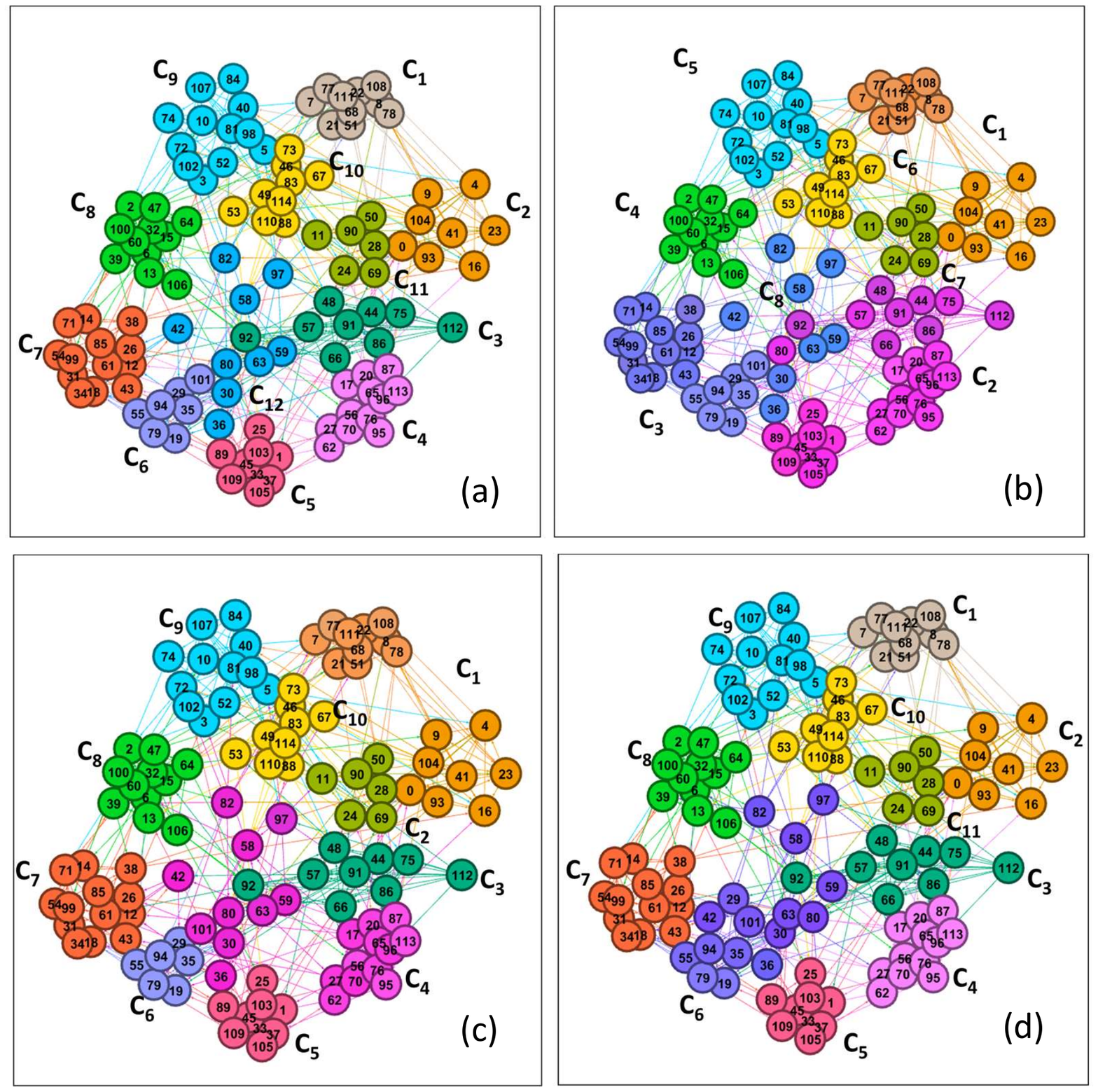}
		\caption{\bf{Visualization of the ground truth communities vs. the identified communities using \method on American College Football Network (ACFN), which has 12 observed players communities i.e. $C \in \{C_1,C_2 \dots C_{12}\}$. \textbf{(a):} Represents the visualization of the network colored with ground truth communities. \textbf{(b):} Shows the visualization of the network colored with predicted communities at time $ \frac{1}{3} T$, where $T$ is total time stamps. \textbf{(c):} Shows the visualization of the network colored with predicted communities at time $ \frac{2}{3} T$. \textbf{(d):} Shows the visualization of the network colored with predicted communities at time $T$. We see that reconstructed views using \method helps to identify the communities changing over time.}}
		\label{fig:realAllcommunities}
	\end{center}
\end{figure*}
\subsection{\method at work}
We construct the ACFN tensor data with the player-player interaction to a 115 x 115 grid, and considering the time as the third dimension of the tensor. Therefore, each element in the tensor is an integer value that represents the number of interactions between players at a specific moment of time. Our aim is to find the players communities (ground truth communities = 12) changed over time in football dataset. In order to evaluate the effectiveness of our method on football dataset, we compare the ground-truth communities against the communities found by the our method. Figure \ref{fig:realAllcommunities} shows a visualization of the football network over time, with nodes colored according to the observed communities. American football leagues are tightly-knit communities because of very limited matches played across communities. Since these communities generally do not overlap, we perform hard clustering. We find that communities are volatile and players belongs to  community {\#}12 (from subfigure (a)) are highly dynamic in forming groups. We observe that \method is able to find relevant communities and also shows the ability to capture the changes in forming those communities in temporal networks.
\hide{
\subsection{Complexity Comparison}
The complexities are given in table \ref{table:complexAnalysis}.

\begin{table*}[h!]
	\small
	\begin{center}
		\begin{tabular}{ |c||c|c|c| }
			\hline
			Methods & Time Complexity & Space Complexity & Reference \\
			\hline
			\method&$O(QSt_{new})$&$pQ(pT+t_{new}+R)+(T+t_{old})R+Q^3$&\\
			OnlineCP&$O(NSt_{new}R)$&$St_{new}+(2T+t_{old})R+(N-1)R^2$&\cite{zhou2016accelerating}\\
			SambaTen&$O(nnz(X)R(N-1)$& $nnz(X)+(T+t_{old}+2S)R+2R^2$&\cite{gujral2017sambaten}\\
			RLST&$O(R^2S)$&$St_{new}+(T+t_{old}+2S)R+2R^2$&\cite{nion2009adaptive}\\
			ParaComp&$O(QS(t_{old}+t_{new}))$&$St_{new}+(T+t_{old})R+pQ^3$&\cite{sidiropoulos2014parallel}\\
			\hline
		\end{tabular}
		\caption{\textbf{Complexity comparison between \method and state-of-art methods.}}
		\label{table:complexAnalysis}
	\end{center}
	\vspace{-0.2in}
\end{table*}
%%%%%%%%%%%%%%%%%%%%%%%%%%%%%%%%%%%%%%%%%%%%%%%%%%%%%%%%%%%%%%%%%%%%%%%%%%%%%%%%%%%%%%%%%%%%%%%%%%% 
}